\documentclass[11pt]{article}

\usepackage[preprint]{acl}

\usepackage{times}
\usepackage{latexsym}
\usepackage{minitoc}
\usepackage{titletoc}

\usepackage[T1]{fontenc}
\usepackage[utf8]{inputenc}

\usepackage{microtype}

\usepackage{inconsolata}

\usepackage{graphicx}

\usepackage{amsmath}
\usepackage{amssymb}
\usepackage{booktabs}
\usepackage{multirow}
\usepackage{tabularx}
\usepackage[most]{tcolorbox}
\usepackage{soul}
\usepackage{bm}
\usepackage{algorithm}

\usepackage[normalem]{ulem}
\usepackage{cleveref}
\usepackage{tikz}
\newcommand*\circled[1]{\tikz[baseline=(char.base)]{
            \node[shape=circle,draw,inner sep=0.4pt] (char) {#1};}}
\usepackage{algpseudocode}
\usepackage{caption}
\usepackage{enumitem}
\usepackage{subcaption}
\usepackage{xcolor}
\usepackage[table,x11names,dvipsnames]{xcolor}

\definecolor{lightgreen}{RGB}{220,245,220}
\newcommand{\hlg}[1]{\colorbox{lightgreen}{#1}}
\usepackage{hyperref}
\definecolor{ChineseRed}{HTML}{9F363A}
\definecolor{EyeProtectGreen}{HTML}{00693e}

\definecolor{BlueViolet}{rgb}{0.54, 0.17, 0.89}

\newcommand{\ours}{\texttt{Doc2Atom}}

\hypersetup{
    colorlinks=true,
    linkcolor=EyeProtectGreen,
    citecolor=EyeProtectGreen,
    filecolor=EyeProtectGreen,      
    urlcolor=EyeProtectGreen,
    }

\title{Doc-to-Atom: Learning to Compile and Compose Memory Atoms}

\author{
\textbf{Xingjian Diao}$^{1,2\dagger}$,
\textbf{Wenbo Li}$^1$,
\textbf{Yashas Malur Saidutta}$^1$,
\textbf{Avinash Amballa}$^1$,
   \\
\textbf{Lazar Valkov$^1$,}
\textbf{Srinivas Chappidi$^1$}
\\
$^1$AI Center-Mountain View, Samsung Electronics
\\
$^2$Dartmouth College
\\
\texttt{xingjian.diao.gr@dartmouth.edu}
\\
\texttt{wenbo.li1@samsung.com}
}
\begin{document}
\maketitle

\newcommand\blfootnote[1]{%
  \begingroup
  \renewcommand\thefootnote{}\footnote{#1}%
  \addtocounter{footnote}{-1}%
  \endgroup
}

\blfootnote{$^\dagger$Work done during an internship at AI Center-Mountain View, Samsung Electronics}

\begin{abstract}
Long input sequences are central to document understanding and multi-step reasoning in Large Language Models, yet the quadratic cost of attention makes inference both memory-intensive and slow. Context distillation mitigates this by compressing contextual information into model parameters, and recent work such as Doc-to-LoRA amortizes context distillation into a single forward pass that generates one LoRA adapter per document. However, producing a single monolithic adapter for all queries leads to irrelevant-query interference, limited compositional recall, and poor scalability to long-document reasoning. To address these challenges, we propose Doc-to-Atom (\ours{}), a compositional parametric memory framework that decomposes each document into semantically typed knowledge atoms. Each atom is compiled into an independent micro-LoRA adapter and a provenance retrieval key. At inference time, a lightweight query router selects and assembles only the relevant atoms into a query-specific adapter, which is then injected into a frozen base model. The entire system is trained end-to-end through a multi-objective distillation framework. Experiments on six diverse QA benchmarks demonstrate that \ours{} outperforms Doc-to-LoRA baselines while reducing the memory cost of document internalization. 

\end{abstract}

\begin{figure}[tb]
\centering
\resizebox{0.48\textwidth}{!}{
\includegraphics{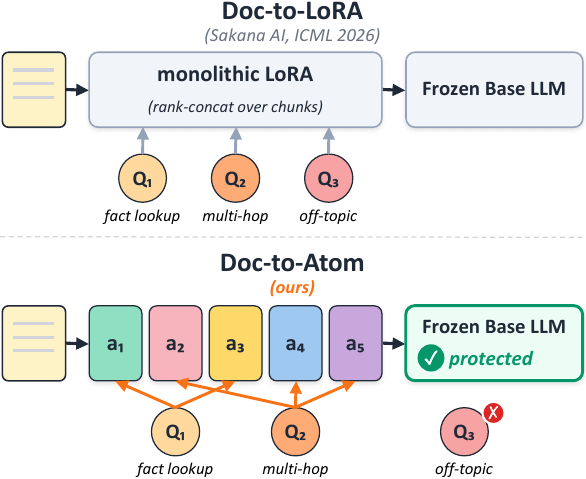}
}
\caption[Doc-to-LoRA vs.\ Doc-to-Atom]{%
\textbf{Doc-to-LoRA vs.\ Doc-to-Atom.}
\emph{Top:} Doc-to-LoRA rank-concatenates fixed-size token chunks into a single document-level LoRA that is applied to all queries, without semantic structure or a per-chunk index. 
\emph{Bottom:} \ours{} decomposes the document into typed semantic atoms; each $a_i$ carries a key, a micro-LoRA, and an optional micro-KV. A two-stage router uses the keys as a per-atom index and assembles only the relevant atoms into a query-specific adapter, with support for refusing queries outside the document.}
\vspace{-0.5cm}
\label{fig:teaser}
\end{figure}

\section{Introduction}
\label{sec:intro}

Large Language Models have achieved remarkable success across natural language understanding tasks by leveraging in-context learning (ICL), where relevant information is placed directly into the model's context window at inference time \citep{brown2020language}. This paradigm allows the model to condition its outputs on arbitrary textual evidence without parameter updates. However, in-context learning suffers from scalability bottlenecks: the quadratic attention complexity causes both latency and memory consumption to grow rapidly with input length \citep{liu2024lost}, and generation quality degrades under longer context lengths due to attention dilution \citep{hsieh2024ruler}. These limitations motivate an alternative paradigm known as context distillation, which compresses contextual information into model parameters so that the model can reproduce context-conditioned behavior without retaining the original text at inference time \citep{askell2021general,snell2023learning}.

While context distillation (CD) eliminates the runtime cost of long contexts, its standard formulation requires per-prompt supervised fine-tuning, which is prohibitively expensive when information sources change frequently \citep{zhang2023instruction}. To overcome this limitation, Doc-to-LoRA (D2L) \citep{charakorn2026doc} proposes to meta-learn the context distillation process by training a hypernetwork \citep{ha2016hypernetworks} that maps any unseen document to a LoRA adapter \citep{hu2022lora} in a single forward pass. This amortizes CD across documents, enabling fast, parameter-efficient knowledge updates without any per-document training. Nevertheless, D2L inherits a fundamental architectural constraint: it produces a single monolithic adapter per document that is applied uniformly to all downstream queries. This design introduces three interrelated problems. \textbf{\circled{1}} When a query is unrelated to the internalized content, the adapter still modifies the model's hidden states, causing
irrelevant-query interference.
\textbf{\circled{2}} For documents containing dozens of heterogeneous facts, a flat low-rank adapter lacks the capacity to selectively surface the evidence required by each query, resulting in limited compositional recall.
\textbf{\circled{3}} As document length grows, the fixed-rank adapter must compress increasingly more content into the same parameter budget, which limits scalability on long-document reasoning where the relevant evidence spans many regions of the text.

To address these challenges, we propose Doc-to-Atom (\ours{}), a framework that fundamentally restructures the document internalization pipeline from one document to one adapter into one document to many composable atoms (as shown in Figure~\ref{fig:teaser}). Our key insight is that \uline{documents should be decomposed into semantically typed knowledge atoms, which are minimal and self-contained units of information}. Each atom is independently compiled into a micro-LoRA adapter, a provenance retrieval key, and an optional compact KV prototype. At query time, a learned two-stage router identifies the small subset of atoms relevant to the current question and assembles their micro-adapters into a query-specific composite LoRA, which is then injected into the frozen base model's designated memory layers. For queries unrelated to the internalized document, the router produces near-zero routing weights, allowing the model to fall back without interference.

Concretely, the \ours{} pipeline operates in two phases. In the offline memory compilation phase, each input document is decomposed into atoms via an LLM-driven semantic annotation process, and then a shared text encoder followed by an atom-level memory compiler produces per-atom micro-LoRA factors, provenance keys, and sparse memory masks. In the online query-conditioned assembly phase, given a query, a query encoder produces a query embedding that is matched against provenance keys via a two-stage router. The top-ranked atoms are selected, their micro-LoRA factors are composed via routing-weight-based summation and gated by sparse masks, and the resulting assembled adapter is injected into the base model for answer generation. Our main contributions are summarized as follows:

\begin{itemize}[leftmargin=*]
    \item We introduce atom-level document decomposition for parametric memory, replacing monolithic document-to-adapter compression with fine-grained, semantically typed knowledge units that enable selective retrieval and composition.
    \item We design a compositional memory compilation and query-conditioned assembly architecture that produces per-atom micro-LoRA adapters, routes queries to relevant atoms, and assembles query-specific adapters dynamically.
    \item We propose a comprehensive multi-objective training framework with curriculum-based scheduling that jointly optimizes memory generation, routing accuracy, irrelevant-query robustness, knowledge protection of the base LLM, and multi-atom composition consistency.
    \item Experiments on six diverse QA benchmarks demonstrate the effectiveness of \ours{} over in-parameter knowledge baselines, offering a step toward compositional parametric memory.
\end{itemize}

\section{Related Work}

\paragraph{Hypernetworks for Language Model Adaptation.}
Hypernetworks \citep{ha2016hypernetworks} generate the parameters of a target network conditioned on some input, and have been increasingly applied to efficient LLM adaptation. HyperDecoders \citep{ivison2022hyperdecoders} produce instance-specific decoder parameters for multi-task NLP, while HINT \citep{ivison2023hint} extends hypernetwork instruction tuning for few-shot generalization. HyperLoRA \citep{lv2024hyperloRA} uses constrained low-rank adapter generation for cross-task transfer. More recently, Text-to-LoRA \citep{charakorn2025text} trains a hypernetwork that directly maps context strings to LoRA parameters. Doc-to-LoRA \citep{charakorn2026doc} builds upon this line by meta-training a hypernetwork with the context distillation objective, enabling zero-shot internalization of unseen documents. While these methods demonstrate the promise of hypernetwork-based adaptation, they all produce monolithic adapters that cannot selectively respond to different queries over the same internalized content. We find that this leads to both degraded long-context recall (Table~\ref{tab:main-results}) and irrelevant-query interference (Table~\ref{tab:irr-f1}).

\begin{figure*}[htbp]
  \centering
  \includegraphics[width=\textwidth]{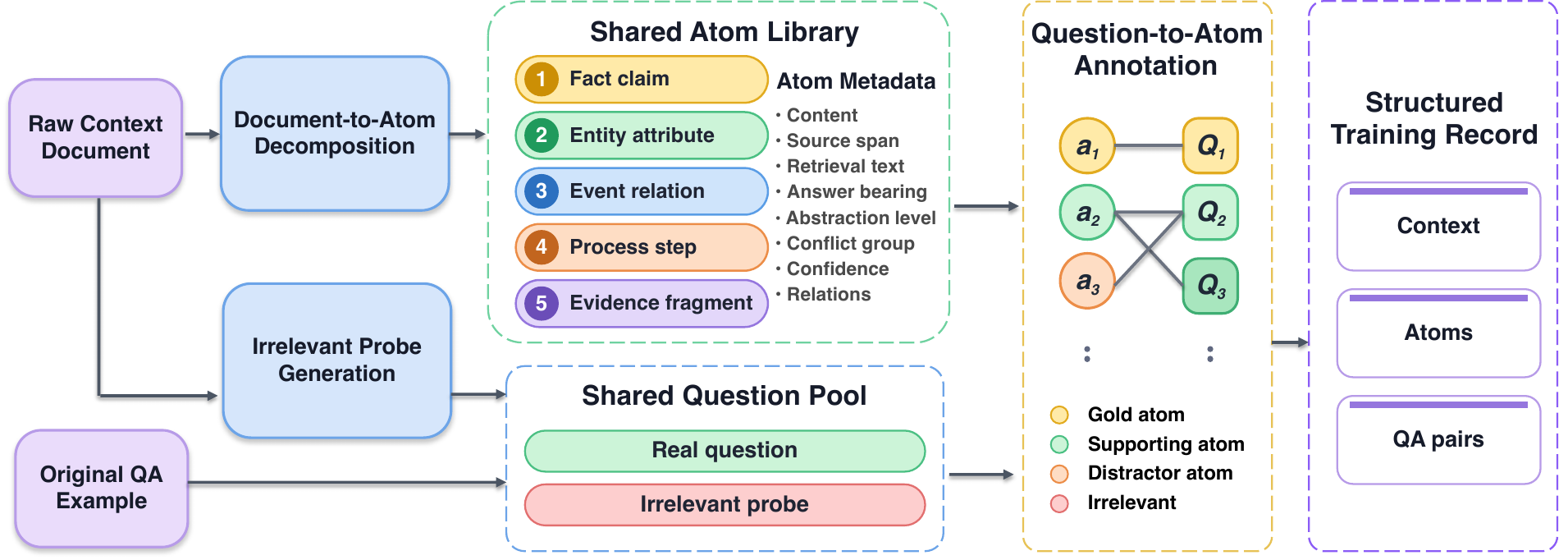}
  \caption{\textbf{\ours{} data atomization and processing pipeline.} Raw context documents are decomposed into semantically typed atoms stored in a shared atom library with rich metadata (e.g., semantic type, retrieval text, and inter-atom relations). Original QA examples are augmented with irrelevant probes to form a shared question pool. The question-to-atom annotation stage aligns each question to relevant atoms, producing structured training records containing context, atoms, and QA pairs with gold, supporting, and distractor labels.}
  \vspace{-0.4cm}
  \label{fig:atomization}
\end{figure*}

\paragraph{Context Distillation and Compression.}
Context distillation trains a model to reproduce context-conditioned behavior without the original context at inference time \citep{askell2021general,snell2023learning}. The core idea of compressing a larger model's knowledge into a smaller one traces back to model compression \citep{bucilua2006model} and knowledge distillation \citep{hinton2015distilling}. Several recent works extend this paradigm: \citet{caccia2025training} train plug-and-play knowledge modules via deep context distillation; Cartridges \citep{eyuboglu2025cartridges} leverage sleep-time compute for prefix-tuning-based context distillation; and Generative Adapter \citep{chen2025generative} optimizes a hypernetwork using next-token prediction. Complementary approaches focus on prompt compression: Gisting \citep{mu2024learning} trains gist tokens that compress task instructions via a prefix-tuning \citep{li2021prefix} parameterization, while LLMLingua-2 \citep{pan2024llmlingua} performs task-agnostic prompt compression via data distillation. MEND \citep{li2024mend} meta-learns demonstration distillation for efficient in-context learning. Our work differs from all these approaches by decomposing documents into atom-level units before compilation, enabling query-conditioned selective assembly rather than monolithic compression.

\paragraph{Knowledge Editing and Protection.}
Parametric knowledge internalization risks overwriting the base model's existing capabilities. AlphaEdit \citep{fang2025alphaedit} proposes null-space constrained knowledge editing that restricts parameter updates to subspaces orthogonal to previously used directions. Sparse memory finetuning \citep{lin2025continual} encourages different knowledge to be written into different parameter subregions, reducing inter-task interference. Our framework incorporates both L2-based \citep{kirkpatrick2017overcoming} and null-space \citep{fang2025alphaedit} knowledge protection mechanisms, combined with learned sparse memory masks that route different atom types to different layer-module combinations, thereby minimizing cross-atom interference during composition.

\paragraph{Long-Context Understanding.}
Handling long documents remains challenging for LLMs due to attention complexity and positional encoding limitations. Lost-in-the-middle phenomena \citep{liu2024lost} and attention noise \citep{ye2025differential} degrade performance on long inputs. Approaches such as activation beacons \citep{zhang2025long} and infinite ICL \citep{cao2025infiniteicl} attempt to extend effective context length, while continual learning with hypernetworks \citep{von2020continual} addresses sequential knowledge accumulation. \ours{} sidesteps these challenges entirely by compiling documents into parametric memory offline, eliminating the need to process long contexts at inference time while preserving fine-grained access through atom-level routing.

\begin{figure*}[tb]
  \centering
  \includegraphics[width=\textwidth]{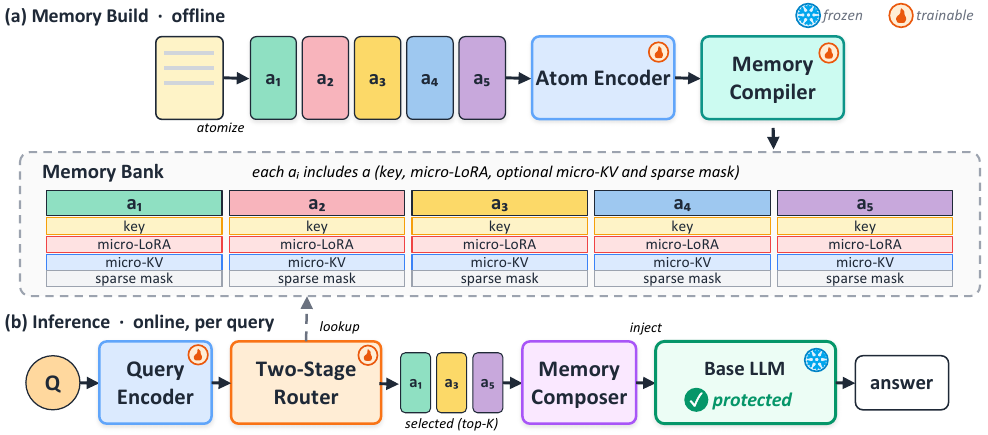}
  \caption{\textbf{\ours{} pipeline.}
  \textbf{(a) Memory build (offline).} The input document is decomposed into typed semantic atoms $a_1,\dots,a_n$. A shared Atom Encoder embeds every atom, and a Memory Compiler produces a per-atom provenance key,
  a micro-LoRA factor, an optional micro-KV prototype and 
  sparse mask, which together populate the Memory Bank.
  \textbf{(b) Inference (online, per query).} A Query Encoder (sharing weights with the Atom Encoder) embeds the query. A two-stage router consults the provenance keys to retrieve the top-$K$ relevant atoms from the bank; the Memory Composer assembles their micro-LoRA factors
  into a query-specific adapter, which is injected into the frozen base LLM (together with the micro-KV prototypes when available) to produce the answer.}
  \vspace{-0.2cm}
  \label{fig:pipeline}
\end{figure*}

\section{Method}
\label{sec:method}

We now present \ours{} in detail. Our goal is to internalize a document
into compositional parametric memory in a way that supports selective
retrieval and query-conditioned composition. To this end, \ours{}
consists of three components: a data atomization pipeline that
decomposes documents into structured knowledge atoms
(\S\ref{sec:method-data}); a compositional memory framework that
compiles atoms and assembles query-specific adapters
(\S\ref{sec:method-framework}); and a multi-objective training scheme
that optimizes the system end-to-end (\S\ref{sec:method-training}).

\subsection{Data Atomization and Processing}
\label{sec:method-data}

The data atomization and processing pipeline, illustrated in Figure~\ref{fig:atomization}, transforms raw QA datasets into structured training records through three sequential stages: document-to-atom decomposition, irrelevant probe generation, and question-to-atom annotation.

\paragraph{Document-to-Atom Decomposition.}
Unlike previous methods which segment documents into fixed-size token chunks regardless of semantic boundaries, \ours{} requires an LLM to explicitly decompose each context document $D$ into a set of non-overlapping, semantically self-contained knowledge atoms $\mathcal{A} = \{a_1, a_2, \dots, a_N\}$. Each atom $a_i$ represents the minimal independently retrievable and writable knowledge unit within the document, drawn from five semantic types: fact claims, entity attributes, event relations, process steps, and evidence fragments. Beyond the semantic type, each atom carries the following metadata:
(i)~\emph{content}, a self-contained text preserving critical qualifiers;
(ii)~\emph{source span}, strictly extracted from the original document;
(iii)~\emph{retrieval text}, used as the indexing anchor;
(iv)~an \emph{answer-bearing} flag indicating whether the atom can directly carry an answer; 
(v)~an \emph{abstraction level} distinguishing abstract, evidence, and hybrid atoms;
(vi)~a \emph{conflict group} identifier marking mutually exclusive atoms;
(vii)~a continuous \emph{confidence} score in $[0,1]$;
and (viii)~a set of \emph{inter-atom relations}.

\paragraph{Irrelevant Probe Generation.}
To train the query router to remain silent on queries unrelated to the internalized content, we synthesize irrelevant probes for each document. These probes are surface-similar questions that mention entities or topics that appear in the document, but whose answers cannot be derived from its content. This forces the router to learn fine-grained boundaries of answerability that go beyond superficial lexical overlap, ensuring that the system falls back to base-model behavior when no relevant atoms exist.

\paragraph{Question-to-Atom Annotation.}
Given the atom library $\mathcal{A}$ and the augmented question pool containing real questions and irrelevant probes, an LLM aligns each question $q$ to atoms by producing a normalized question type, a set of gold atom IDs $\mathcal{A}_{\text{gold}}$ that constitute the minimal atom set required to answer the question, supporting atom IDs $\mathcal{A}_{\text{supp}}$ that aid reasoning but are not strictly necessary, and distractor atom IDs $\mathcal{A}_{\text{dist}}$ that are surface-similar and most likely to be falsely activated. Boolean flags for irrelevance and conflict are also generated.

\subsection{Compositional Memory Framework}
\label{sec:method-framework}

The \ours{} training pipeline, depicted in Figure~\ref{fig:pipeline}, operates through offline memory compilation and online query-conditioned assembly, connected by a teacher--student training objective.

\paragraph{Shared Text Encoder.}
Rather than introducing a separate encoding model, \ours{} reuses the first $L_{\text{enc}}$ layers of the frozen base model as a shared text encoder for both atoms and queries. Each atom's textual content is tokenized and fed through these layers, with an early-exit mechanism that extracts hidden states at a designated intermediate layer. A learned projection head maps the pooled hidden state to a $d$-dimensional atom embedding $e_i \in \mathbb{R}^d$. This design avoids introducing additional model parameters for encoding while leveraging the base model's pre-trained representations.

\paragraph{Atom-Level Memory Compiler.}
The memory compiler takes atom embeddings $e_i$ as input and produces
four types of outputs through a shared trunk MLP (LayerNorm followed
by two linear layers with a GELU activation in between) and four
parallel heads. First, a provenance key head (LayerNorm followed by a
linear projection) maps each atom to a retrieval anchor vector
$k_i \in \mathbb{R}^{d_k}$ that serves as the index for routing.
Second, micro-LoRA heads produce per-atom low-rank factors
$A_{m,i}^{(\ell)} \in \mathbb{R}^{r \times d_{\text{in}}}$ and
$B_{m,i}^{(\ell)} \in \mathbb{R}^{d_{\text{out}} \times r}$ for each
target module $m$ and each designated memory layer $\ell$, with $B$
initialized to zero to ensure the memory produces no perturbation at
the start of training. Third, an optional micro-KV head generates
compact key-value prototypes per atom for preserving local evidence
and sequential information. Fourth, an optional sparse mask head
outputs per-atom logits that, after aggregation via routing weights
and sigmoid activation, form a sparse gate $g \in [0,1]^{n_L \times n_M}$
controlling which layer-module combinations each atom's knowledge is written to.

\paragraph{Two-Stage Query Router.}
Given a query embedding $e_q \in \mathbb{R}^d$, the router identifies relevant atoms through two stages. In the first stage, the query is projected and L2-normalized to obtain $k_q \in \mathbb{R}^{d_k}$, then matched against all provenance keys $k_i$ via cosine maximum inner product search to retrieve the top candidates. In the optional second stage, these candidates are reranked by a frozen cross-encoder that scores the query and atom text pairs. A learned metadata bias incorporating atom confidence, answer-bearing status, conflict group membership, and semantic type is added to inject structural priors from the annotation stage. The final score $s_i$ combines the similarity, optional reranking, and metadata bias, from which the top-$K$ atoms are selected and their scores are softmax-normalized into sparse routing weights $w_i$.

\paragraph{Memory Composer and Injection.}
The memory composer assembles per-atom parameters into a query-specific adapter. For each layer $\ell$ and module $m$, micro-LoRA factors are combined via routing-weight-based summation:
\begin{equation}
    A_c^{(\ell,m)} = \sum_{i=1}^K w_i A_{m,i}^{(\ell)}, \quad B_c^{(\ell,m)} = \sum_{i=1}^K w_i B_{m,i}^{(\ell)}.
\end{equation}

The composed factors are then element-wise multiplied by the aggregated sparse gate $g$ to enforce selective writing. When micro-KV is enabled, selected atoms' KV slots are scaled by routing weights and concatenated along the sequence dimension as a compact prefix. All writes are restricted to the last $n_L$ decoder layers, following the principle that memory should be injected close to the output to minimize interference with lower-level linguistic processing.

\subsection{Training Objective}
\label{sec:method-training}

The training follows a teacher--student paradigm. The teacher forward pass feeds the frozen base model with the full context $D$ concatenated with the query $q$, producing full-context reference logits $y_{\text{teacher}}$. The student forward pass feeds the same frozen model with only the query $q$, but with the assembled composite LoRA injected into the designated memory layers, producing student logits $y_{\text{student}}$. The total training loss $\mathcal{L}_{\text{total}}$ is a weighted sum of several objectives designed to jointly optimize memory generation, routing accuracy, irrelevant-query robustness, and multi-atom composition consistency.

\paragraph{Language Modeling and Distillation.}
The fundamental objective is to reproduce the teacher's answer distribution purely through the parametric memory assembled from selected atoms. We apply standard cross-entropy loss on the answer tokens:
\begin{equation}
    \mathcal{L}_{\text{LM}} = -\sum_{t} \log P_{\text{student}}(y_t \mid y_{<t}, q, \mathcal{A}).
\end{equation}

We also apply a temperature-scaled Kullback-Leibler divergence between student and teacher logits at answer positions to distill the teacher's knowledge:
\begin{equation}
    \mathcal{L}_{\text{distill}} = \text{KL}\left( \sigma(y_{\text{teacher}} / \tau) \parallel \sigma(y_{\text{student}} / \tau) \right),
\end{equation}
where $\sigma$ denotes the softmax function and $\tau$ is the temperature parameter.

\paragraph{Routing Supervision.}
To train the router to select the correct atoms, we apply a per-atom weighted binary cross-entropy loss between the routing scores $s_i$ and the gold atom masks $m_i \in \{0, 1\}$:
\begin{multline}
    \mathcal{L}_{\text{route}} = -\sum_{i} \alpha_i \big[ m_i \log(\sigma(s_i)) \\
    + (1 - m_i) \log(1 - \sigma(s_i)) \big],
\end{multline}
where $\alpha_i$ takes values $\{0.5, 1.0, 1.5\}$ for negative, gold, and distractor atoms respectively, so distractor false positives are
penalized more heavily than ordinary negatives. When the second stage is enabled, we apply a listwise softmax cross-entropy loss
$\mathcal{L}_{\text{rerank}}$ for the reranking scores.

\paragraph{Irrelevant Query Suppression.}
For queries that are unrelated to the internalized document, the router should produce near-zero routing weights, and the composed adapter should not perturb the base model. We enforce this via a norm penalty on the composed LoRA factors for irrelevant queries:
\begin{equation}
    \mathcal{L}_{\text{irrel}} = \mathbb{E}_{q \sim \mathcal{Q}_{\text{irrel}}} \sum_{\ell, m} \left( \|A_c^{(\ell,m)}\|_2^2 + \|B_c^{(\ell,m)}\|_2^2 \right).
\end{equation}

\paragraph{Knowledge Protection.}
To prevent the assembled adapter from overwriting the base model's existing capabilities, we apply an L2 constraint on the composed deltas:
\begin{equation}
    \mathcal{L}_{\text{protect}} = \sum_{\ell, m} \|\Delta W^{(\ell,m)}\|_F^2,
\end{equation}
where $\Delta W^{(\ell,m)} = B_c^{(\ell,m)} A_c^{(\ell,m)}$. Alternatively, a null-space projection penalty can be used to restrict updates to directions orthogonal to the base model's principal components.

\paragraph{Sparse Regularization.}
We encourage the sparse gates $g$ to activate only a subset of layer-module combinations, pushing activation rates toward a target density $\rho$:
\begin{equation}
    \mathcal{L}_{\text{sparse}} = \sum_{\ell, m} \left( \mathbb{E}[g_{\ell,m}] - \rho \right)^2,
\end{equation}
where $\mathbb{E}[g_{\ell,m}]$ is the mean gate value at $(\ell, m)$ averaged over valid atoms in the batch.

\paragraph{Composition Consistency.}
To ensure that combining multiple atoms does not degrade performance compared to using single atoms, we enforce a symmetric KL divergence between the main forward pass and a forced-routing forward pass using gold masks:
\begin{equation}
    \mathcal{L}_{\text{composition}} = \text{KL}_{\text{sym}}\left( y_{\text{student}} \parallel y_{\text{forced}} \right).
\end{equation}

Additional robustness losses include $\mathcal{L}_{\text{conflict}}$, an intra-group listwise cross-entropy requiring the router to activate only the gold atom within each conflict group, and $\mathcal{L}_{\text{confidence}}$, which aligns routing weights with annotation-side confidence priors.

Training proceeds through a multi-stage curriculum to reduce early coupling conflicts. The curriculum progressively increases the number of retrieved atoms $K$, activates robustness and consistency losses, and strengthens sparse writing regularization. The detailed configuration of the hyperparameters and curriculum stages is provided in the Appendix 
\S\ref{app:impl} and \S\ref{app:curriculum}.

\begin{table*}[tb]
\centering
\footnotesize
\setlength{\tabcolsep}{3.2pt}
\renewcommand{\arraystretch}{1.05}
\resizebox{0.95\textwidth}{!}{%
\begin{tabular}{l ccccc ccccc}
\toprule
& \multicolumn{5}{c}{\textbf{Gemma-2-2B-It}}
& \multicolumn{5}{c}{\textbf{Qwen3-4B-Instruct}} \\
\cmidrule(lr){2-6} \cmidrule(lr){7-11}
Dataset
 &  Base$+$ctx & D2L$_{\text{ckpt}}$ & D2L$_{\text{raw}}$ & D2L$_{\text{atom}}$ & \ours{}
 &  Base$+$ctx & D2L$_{\text{ckpt}}$ & D2L$_{\text{raw}}$ & D2L$_{\text{atom}}$ & \ours{} \\
\midrule
\multicolumn{11}{c}{\textit{F1}} \\
\midrule
2WikiMultiHopQA \citep{ho2020constructing}      & 37.69 & 36.08 & 49.84 & 50.39 & 39.62 & 45.31 & 11.25 & 46.06 & 47.43 & 31.74 \\
DROP \citep{dua2019drop}                    & 42.61 & 29.62 & 26.98 & 29.52 & \hlg{28.31} & 45.18 & 10.07 & 27.12 & 30.66 & 27.05 \\
QASPER \citep{dasigi2021dataset}            & 48.62 & 26.34 & 16.46 & 18.39 & \hlg{30.78} & 52.76 & 27.03 & 19.17 & 20.07 & \hlg{31.03} \\
ROPES \citep{lin2019reasoning}              & 71.90 & 65.90 & 52.94 & 53.93 & \hlg{53.52} & 77.63 & 15.26 & 53.06 & 51.95 & \hlg{58.08} \\
SQuAD \citep{rajpurkar2016squad}            & 86.35 & 73.20 & 18.67 & 19.36 & \hlg{59.71} & 78.15 & 29.60 & 15.88 & 18.71 & \hlg{59.82} \\
\cmidrule(lr){1-11}
LongBench-2WikiMQA    & 32.27 & 27.93 & 34.63 & 36.48 & 29.63 & 37.47 & \phantom{0}9.09 & 34.87 & 36.14 & 26.35 \\
LongBench-HotpotQA    & 41.00 & 26.79 & 22.30 & 19.49 & \hlg{27.96} & 52.15 & 11.04 & 25.88 & 25.43 & 24.77 \\
LongBench-MFQA-en     & 39.41 & 20.92 & \phantom{0}7.14 & \phantom{0}4.68 & \hlg{26.00} & 47.18 & 26.69 & 10.70 & \phantom{0}9.73 & \hlg{34.29} \\
LongBench-MFQA-zh     & \phantom{0}7.73 & \phantom{0}2.05 & \phantom{0}1.57 & \phantom{0}0.41 & \hlg{\phantom{0}8.57} & 14.79 & \phantom{0}5.17 & \phantom{0}0.70 & \phantom{0}0.54 & \hlg{\phantom{0}9.27} \\
LongBench-MuSiQue     & 21.62 & \phantom{0}9.14 & \phantom{0}6.95 & \phantom{0}7.39 & \hlg{20.86} & 18.14 & \phantom{0}5.24 & 12.25 & \phantom{0}6.41 & \hlg{22.32} \\
LongBench-NarrativeQA & 22.67 & 13.31 & \phantom{0}4.64 & \phantom{0}6.21 & \hlg{20.58} & 27.40 & 11.67 & \phantom{0}2.83 & \phantom{0}3.46 & \hlg{17.87} \\
LongBench-QASPER      & 37.05 & 19.81 & 12.29 & 17.10 & \hlg{29.85} & 45.14 & 21.71 & 14.84 & 14.56 & \hlg{31.80} \\
LongBench-TriviaQA    & 86.67 & 82.14 & 83.27 & 83.48 & 79.54 & 85.16 & 50.81 & 85.99 & 85.85 & 47.77 \\
\cmidrule(lr){1-11}
\textit{Overall}      & 48.79 & 37.93 & 28.55 & 29.41 & \colorbox{lightgreen}{\textbf{37.99}} & 52.22 & 18.70 & 28.85 & 29.30 & \colorbox{lightgreen}{\textbf{35.72}} \\
\midrule
\multicolumn{11}{c}{\textit{ROUGE-L}} \\
\midrule
2WikiMultiHopQA \citep{ho2020constructing}      & 37.69 & 36.08 & 49.79 & 50.25 & 39.62 & 45.27 & 11.17 & 46.03 & 47.43 & 31.74 \\
DROP \citep{dua2019drop}                    & 42.61 & 29.55 & 26.98 & 29.52 & \hlg{28.31} & 45.18 & 10.02 & 27.12 & 30.66 & 27.05 \\
QASPER \citep{dasigi2021dataset}            & 47.26 & 25.39 & 16.25 & 18.19 & \hlg{30.44} & 50.72 & 25.39 & 18.91 & 19.83 & \hlg{30.70} \\
ROPES \citep{lin2019reasoning}              & 71.90 & 65.85 & 52.94 & 53.93 & \hlg{53.52} & 77.63 & 15.25 & 52.98 & 51.95 & \hlg{58.08} \\
SQuAD \citep{rajpurkar2016squad}            & 86.20 & 72.96 & 18.57 & 19.21 & \hlg{59.71} & 77.89 & 29.04 & 15.67 & 18.48 & \hlg{59.66} \\
\cmidrule(lr){1-11}
LongBench-2WikiMQA    & 32.27 & 27.86 & 34.57 & 36.48 & 29.63 & 37.47 & \phantom{0}8.92 & 34.87 & 36.08 & 26.24 \\
LongBench-HotpotQA    & 40.78 & 26.37 & 22.21 & 19.49 & \hlg{27.82} & 51.78 & 10.50 & 25.88 & 25.24 & 24.67 \\
LongBench-MFQA-en     & 38.31 & 19.70 & \phantom{0}6.91 & \phantom{0}4.68 & \hlg{25.83} & 45.60 & 24.31 & \phantom{0}9.92 & \phantom{0}8.94 & \hlg{33.31} \\
LongBench-MFQA-zh     & \phantom{0}7.73 & \phantom{0}2.05 & \phantom{0}1.37 & \phantom{0}0.41 & \hlg{\phantom{0}8.57} & 14.79 & \phantom{0}5.17 & \phantom{0}0.70 & \phantom{0}0.54 & \hlg{\phantom{0}9.27} \\
LongBench-MuSiQue     & 21.62 & \phantom{0}9.14 & \phantom{0}6.95 & \phantom{0}7.39 & \hlg{20.86} & 18.05 & \phantom{0}5.00 & 12.25 & \phantom{0}6.41 & \hlg{22.27} \\
LongBench-NarrativeQA & 22.37 & 13.27 & \phantom{0}4.64 & \phantom{0}6.15 & \hlg{20.27} & 26.56 & 11.20 & \phantom{0}2.83 & \phantom{0}3.46 & \hlg{17.82} \\
LongBench-QASPER      & 35.58 & 18.87 & 11.89 & 16.68 & \hlg{29.26} & 42.93 & 20.35 & 14.64 & 14.12 & \hlg{31.46} \\
LongBench-TriviaQA    & 86.67 & 82.14 & 83.27 & 83.48 & 79.54 & 85.14 & 50.53 & 85.99 & 85.85 & 47.74 \\
\cmidrule(lr){1-11}
\textit{Overall}      & 48.43 & 37.62 & 28.45 & 29.32 & \colorbox{lightgreen}{\textbf{37.88}} & 51.65 & 18.13 & 28.74 & 29.17 & \colorbox{lightgreen}{\textbf{35.58}} \\
\bottomrule
\end{tabular}%
}
\caption{\textbf{Per-dataset F1 and ROUGE-L scores (\%).}
We compare \ours{} against three D2L baselines on two frozen base LLMs.
\textbf{D2L$_{\text{ckpt}}$} uses the official checkpoint released by
Sakana AI~\citep{charakorn2026doc} (trained on FineWeb-Edu~\citep{lozhkov2024fineweb-edu},
PwC~\citep{chevalier2023adapting}, SQuAD, ROPES, and DROP);
\textbf{D2L$_{\text{raw}}$} retrains D2L on our five source datasets;
and \textbf{D2L$_{\text{atom}}$} retrains D2L on the atomized version
of these datasets. \textbf{Base$+$ctx} is an ICL reference with full
document access at inference time. The first five datasets are
in-domain; the LongBench datasets~\citep{bai2024longbench} are evaluated
zero-shot for all methods. \colorbox{lightgreen}{Green} marks cases
where \ours{} beats \textbf{D2L$_{\text{raw}}$}. \textbf{Bold} in the
\textit{Overall} row marks the best parameterized method.}
\vspace{-0.5cm}
\label{tab:main-results}
\end{table*}

\section{Experiments}
\label{sec:exp}

\ours{} addresses limitations of monolithic document-to-adapter internalization that prior parametric methods do not handle (\S\ref{sec:intro}). First, we compare \ours{} against three Doc-to-LoRA (D2L) baselines for accuracy and long-context generalization (Table~\ref{tab:main-results}). Second, we evaluate
refusal on out-of-document queries (Table~\ref{tab:irr-f1}). We then assess per-update compile latency and memory (Table~\ref{tab:efficiency}).

\subsection{Experimental Setup}

\paragraph{Models.} 
We evaluate \ours{} on two frozen instruction-tuned base LLMs from the Doc-to-LoRA setting: Gemma-2-2B-It \citep{team2024gemma} and Qwen3-4B-Instruct-2507 \citep{yang2025qwen3}. Atomization is performed offline by MiniMax-M2.5 \citep{MinimaxM25}.

\paragraph{Baselines.}
On each base LLM, we compare \ours{} against three D2L 
baselines and an in-context reference. 
\textbf{D2L\textsubscript{ckpt}} uses the official Doc-to-LoRA
checkpoint \citep{charakorn2026doc} as-is.
\textbf{D2L\textsubscript{raw}} retrains Doc-to-LoRA from scratch on
our source datasets.
\textbf{D2L\textsubscript{atom}} retrains the same architecture on
the atomized version of these datasets, isolating atom-level
decomposition. \textbf{Base$+$ctx} feeds the full document to the frozen base LLM
at inference, serving as an ICL upper bound.

\paragraph{Tasks and Benchmarks.}
We evaluate on six QA benchmarks across three task categories:
short-context reading comprehension (SQuAD \citep{rajpurkar2016squad},
DROP \citep{dua2019drop}, ROPES \citep{lin2019reasoning}),
long-document and multi-hop QA (2WikiMultiHopQA
\citep{ho2020constructing}, QASPER \citep{dasigi2021dataset}), and
long-context understanding via eight LongBench
\citep{bai2024longbench} subsets. The first two categories are used
for both training and in-distribution testing; LongBench is evaluated
zero-shot for long-context generalization. We report word-level F1
and ROUGE-L as primary metrics.

\paragraph{Configuration.} All other details are in Appendix.

\subsection{Main Results}
\label{sec:exp-main}

\paragraph{\ours{} achieves the best overall accuracy.}
On Gemma-2-2B-It, \ours{} attains the best overall accuracy among all parametric methods. \ours{} reaches $37.99$ F1 and $37.88$ ROUGE-L
overall, outperforming the D2L baseline
(D2L\textsubscript{atom}) by $8.58$ and $8.56$ points. On
Qwen3-4B-Instruct, \ours{} reaches $35.72$ F1 and $35.58$ ROUGE-L,
exceeding D2L\textsubscript{atom} by $6.42$ and $6.41$ points. The
gains are most pronounced on the eight LongBench subsets, evaluated
\emph{zero-shot} for all methods: \ours{} achieves up to $6\times$
higher F1 than D2L\textsubscript{raw} on MFQA, MuSiQue, NarrativeQA,
and QASPER, substantially narrowing the gap to the ICL upper bound
(Base$+$ctx) on tasks unseen during training.

\paragraph{Atomization alone improves D2L.}
Training the same D2L architecture on atomized documents instead of raw documents  (D2L\textsubscript{raw}$\to$D2L\textsubscript{atom}) yields consistent overall gains on both base LLMs ($+0.86$ on Gemma, $+0.45$ on Qwen3), confirming the effectiveness of our atomization pipeline.

\begin{table}[ht]
\centering
\small
\setlength{\tabcolsep}{5pt}
\renewcommand{\arraystretch}{1.05}
\resizebox{\columnwidth}{!}{%
\begin{tabular}{l cccc}
\toprule
Dataset & D2L$_{\text{ckpt}}$ & D2L$_{\text{raw}}$ & D2L$_{\text{atom}}$ & \ours{} \\
\midrule
\multicolumn{5}{l}{\textbf{Gemma-2-2B-It}} \\
\midrule
2WikiMultiHopQA           & \phantom{0}8.53 & 11.31 & 16.40 & \textbf{90.01} \\
DROP                  & \phantom{0}3.14 & \phantom{0}4.48 & \phantom{0}4.04 & \textbf{92.03} \\
QASPER                & \phantom{0}1.20 & 23.49 & 25.67 & \textbf{58.05} \\
ROPES                 & 11.19           & \phantom{0}5.24 & \phantom{0}5.48 & \textbf{94.66} \\
SQuAD                 & \phantom{0}7.61 & \phantom{0}1.52 & \phantom{0}0.87 & \textbf{85.25} \\
\cmidrule(lr){1-5}
LongBench-2WikiMQA    & \phantom{0}8.50 & 15.83 & 22.00 & \textbf{97.50} \\
LongBench-HotpotQA    & \phantom{0}8.00 & 16.00 & 15.83 & \textbf{90.17} \\
LongBench-MFQA-en     & \phantom{0}2.67 & 18.00 & 11.33 & \textbf{95.22} \\
LongBench-MFQA-zh     & 14.50           & 40.50 & 11.50 & \textbf{95.95} \\
LongBench-MuSiQue     & \phantom{0}3.00 & \phantom{0}8.50 & 12.50 & \textbf{95.33} \\
LongBench-NarrativeQA & \phantom{0}1.00 & \phantom{0}7.00 & 24.50 & \textbf{84.33} \\
LongBench-QASPER      & \phantom{0}1.50 & 21.15 & 22.33 & \textbf{61.50} \\
LongBench-TriviaQA    & \phantom{0}1.00 & \phantom{0}8.50 & \phantom{0}5.50 & \textbf{64.00} \\
\midrule
\multicolumn{5}{l}{\textbf{Qwen3-4B-Instruct}} \\
\midrule
2WikiMultiHopQA           & 17.26 & 10.12 & \phantom{0}2.98 & \textbf{18.88} \\
DROP                  & 34.80 & \phantom{0}4.93 & \phantom{0}1.79 & \textbf{54.71} \\
QASPER                & 43.06 & 54.93 & 39.59 & \textbf{90.68} \\
ROPES                 & 50.07 & 12.14 & \phantom{0}5.48 & \textbf{76.94} \\
SQuAD                 & 42.60 & \phantom{0}2.83 & \phantom{0}0.65 & \textbf{54.13} \\
\cmidrule(lr){1-5}
LongBench-2WikiMQA    & 13.20           & 11.00 & \phantom{0}7.50 & \textbf{17.64} \\
LongBench-HotpotQA    & 15.28           & 10.50 & 10.50 & \textbf{19.08} \\
LongBench-MFQA-en     & 17.96           & 17.33 & 20.67 & \textbf{64.00} \\
LongBench-MFQA-zh     & 21.47           & 33.00 & 21.00 & \textbf{95.00} \\
LongBench-MuSiQue     & 14.38           & \phantom{0}6.50 & \phantom{0}6.50 & \textbf{28.58} \\
LongBench-NarrativeQA & 75.04           & \phantom{0}5.50 & \phantom{0}7.00 & \textbf{92.50} \\
LongBench-QASPER      & 38.35           & 50.50 & 40.33 & \textbf{93.25} \\
LongBench-TriviaQA    & \phantom{0}6.54 & \phantom{0}4.00 & \phantom{0}3.00 & \textbf{20.00} \\
\bottomrule
\end{tabular}%
}
\vspace{-0.2cm}
\caption{\textbf{F1 scores (\%) of refusal on irrelevant queries.}
Each document is paired with unrelated questions and the model is prompted to reply \texttt{unanswerable} when the answer cannot be grounded in the context; we report F1 against this target, so higher means more reliable refusal. The first five datasets are in-domain; LongBench rows are held out at training time. Best per row in \textbf{bold}.}
\vspace{-0.5cm}
\label{tab:irr-f1}
\end{table}

\subsection{Refusal on Irrelevant Queries}
For each document we pair it with unrelated questions and prompt the
model to reply \texttt{unanswerable}; Table~\ref{tab:irr-f1} reports
the F1 score against this refusal target. On Gemma-2-2B-It,
\ours{} consistently outperforms all three D2L baselines on every
benchmark, with refusal F1 typically above $85\%$ and peaking at
$97.50$ on LongBench-2WikiMQA, while D2L baselines stay below $40\%$
on most benchmarks. On Qwen3-4B-Instruct, absolute refusal is harder
for every method (Qwen3 tends to attempt an answer even when
unsupported), yet \ours{} still wins every benchmark, reaching QASPER ($90.68$), and
ROPES ($76.94$). This advantage carries over to the eight held-out LongBench subsets,
where \ours{} wins every subset under both base LLMs, confirming that the learned refusal generalizes beyond the training distribution.

\subsection{Efficiency Analysis}
Table~\ref{tab:efficiency} reports the mean per-update compile cost
across the same six benchmarks as Table~\ref{tab:main-results}.

\begin{table}[ht]
\centering
\small
\setlength{\tabcolsep}{6pt}
\renewcommand{\arraystretch}{1.10}
\begin{tabular}{l cc cc}
\toprule
& \multicolumn{2}{c}{\textbf{Gemma-2-2B-It}} & \multicolumn{2}{c}{\textbf{Qwen3-4B-Instruct}} \\
\cmidrule(lr){2-3}\cmidrule(lr){4-5}
Method & Latency & Memory & Latency & Memory \\
\midrule
D2L$_{\text{ckpt}}$ & \textbf{0.31} & 24.02          & \textbf{0.51} & 34.62 \\
\ours{} & 2.10          & \textbf{13.43} & 1.10          & \textbf{\phantom{0}5.02} \\
\bottomrule
\end{tabular}
\caption{\textbf{Efficiency.}
Mean update latency (s) and mean additional GPU memory (GB) of the
compile / internalization step, averaged across the same datasets
as Table~\ref{tab:main-results}. Lower is better ($\downarrow$); best per column
in \textbf{bold}.}
\vspace{-0.4cm}
\label{tab:efficiency}
\end{table}

\paragraph{Additional Memory.}
\ours{} requires substantially less additional GPU memory than D2L: $13.43$\,GB vs.\ $24.02$\,GB on Gemma-2-2B-It (a $44\%$ reduction) and $5.02$\,GB vs.\ $34.62$\,GB on Qwen3-4B (an $85\%$ reduction).
The top-$K$ atom budget bounds the activation footprint by the number of selected atoms, independent of document length, whereas D2L's monolithic adapter must accommodate every chunk and scales with document size.

\paragraph{Update Latency.}
\ours{} is slower per update than D2L ($2.10$\,s vs.\ $0.31$\,s on Gemma-2-2B-It; $1.10$\,s vs.\ $0.51$\,s on Qwen3-4B-Instruct) because the atom encoder, memory compiler, and two-stage router each add a
sequential pass. Since this update is performed offline at document internalization time and decoupled from per-query inference, update latency is not a primary efficiency bottleneck.

\section{Conclusion}
We presented \ours{}, a framework that internalizes each document as a compositional library of memory atoms and retrieves and composes the relevant atoms into a query-specific memory for each query. This
design addresses three limitations of monolithic document-to-adapter compression: (i) irrelevant-query interference, (ii) limited compositional recall, and (iii) poor scalability to long-document reasoning. Across six diverse QA benchmarks, \ours{} achieves higher accuracy than Doc-to-LoRA baselines, more reliably refuses irrelevant queries, and incurs lower internalization memory, offering a step toward
scalable, compositional, and selective parametric memory.

\section{Broader Outlook}
We view \ours{} as a step toward treating document internalization as a core mechanism for knowledge acquisition. Today, LLMs absorb knowledge by repeatedly processing raw text, either through large-scale pretraining or long-context inference, which scales poorly as knowledge grows. If documents can instead be reliably internalized into compact, composable parametric memory, knowledge is compiled once and offline rather than re-processed on every forward pass.

This decoupling could reshape the economics of scale. A model that faithfully internalizes external documents need not spend its context window or pretraining budget memorizing facts that can be compiled on demand, freeing resources for reasoning and generalization. Realizing this vision still requires closing the gap to in-context learning and scaling to far larger models, but compositional parametric memory offers a concrete path forward.

\section*{Limitations}
While \ours{} demonstrates the effectiveness of internalizing documents into compositional parametric memory across a wide range of QA and long-context benchmarks, by decomposing each document into typed knowledge atoms and composing the relevant ones into a query-specific adapter, several limitations remain.

First, our current implementation relies on a fixed semantic taxonomy and a predefined decomposition strategy when atomizing each document. While this design produces stable and interpretable memory units across datasets, different tasks or domains may require finer or coarser atom granularities, or altogether more flexible decomposition schemes. Investigating adaptive, task-specific atom structures is a promising avenue for future work.

Second, our experiments center on relatively small frozen base language models, consistent with recent context distillation studies such as Doc-to-LoRA. Scaling to substantially larger models would demand far greater computational resources than the 8-GPU setup used in this work. While this setting already demonstrates the effectiveness and efficiency of compositional parametric memory under a constrained compute budget, extending \ours{} to frontier-scale models and broader multimodal settings is left for future work.

Third, although \ours{} outperforms the existing state-of-the-art document-internalization method Doc-to-LoRA on a range of datasets, a non-trivial gap to in-context learning still remains. Our work is intended to advance this direction rather than to close the gap entirely, and we believe substantial headroom can be unlocked through better-designed atom content, more effective routing strategies, and a more principled choice of what information should be internalized into the micro-KV memory. We leave a deeper exploration of these directions to future work.

\section*{Ethical Considerations}
All experiments presented in this study were conducted using publicly available datasets and models licensed for academic research purposes. To the best of our knowledge, this work does not present any ethical concerns.

\bibliography{main}

% \vspace{2cm}

\appendix

\startcontents[appendices]
\section*{Contents of Appendix}
\printcontents[appendices]{}{1}{\normalsize}

\section{Benchmarks}
\label{dataset}

\begin{itemize}[leftmargin=*]

\item {
\textbf{SQuAD} \citep{rajpurkar2016squad}:
SQuAD is a large-scale reading comprehension benchmark constructed from Wikipedia articles. Each example consists of a paragraph, a question written by crowdworkers, and an answer that appears as a contiguous span in the corresponding passage. The dataset evaluates whether models can locate and extract the correct answer from a given context rather than selecting from predefined answer choices.
}

\item {
\textbf{DROP} \citep{dua2019drop}:
DROP is a reading comprehension benchmark requiring discrete reasoning over paragraphs. It contains over 96K questions crowdsourced from Wikipedia, where models must perform operations such as addition, counting, sorting, and comparison, going beyond span extraction to test numerical and compositional reasoning abilities.
}

\item {
\textbf{ROPES} \citep{lin2019reasoning}:
ROPES is a reading comprehension benchmark focusing on reasoning over paragraph effects in situations. It provides a background passage describing causal relationships, a novel situation, and questions that require applying the inferred cause--effect relations to the new context, testing models’ ability to transfer knowledge beyond the original text.
}

\item {
\textbf{2WikiMultihop} \citep{ho2020constructing}:
2WikiMultihop is a multi-hop question answering dataset constructed from Wikipedia and Wikidata, designed to require reasoning over multiple paragraphs. Each example includes not only the answer and supporting sentences, but also structured evidence in the form of triples that explicitly capture the reasoning path from the question to the answer, enabling more comprehensive evaluation of multi-hop reasoning.
}

\item {
\textbf{QASPER} \citep{dasigi2021dataset}:
QASPER is an information-seeking question answering benchmark built on full-text NLP research papers. Questions are written after annotators read only the title and abstract, while answers are derived from the full paper with supporting evidence. This setting requires document-level understanding and often involves reasoning over multiple paragraphs, tables, or figures.
}

\item {
\textbf{LongBench} \citep{bai2024longbench}:
LongBench is a bilingual, multi-task benchmark designed to evaluate long-context understanding in large language models. It consists of diverse datasets and task categories, including single-document QA, multi-document QA, summarization, few-shot learning, synthetic tasks, and code completion, with context lengths extending to thousands of tokens, enabling comprehensive evaluation of long-range reasoning and comprehension.
}

\end{itemize}

\begin{figure*}[tb]
\centering
\includegraphics[width=\textwidth]{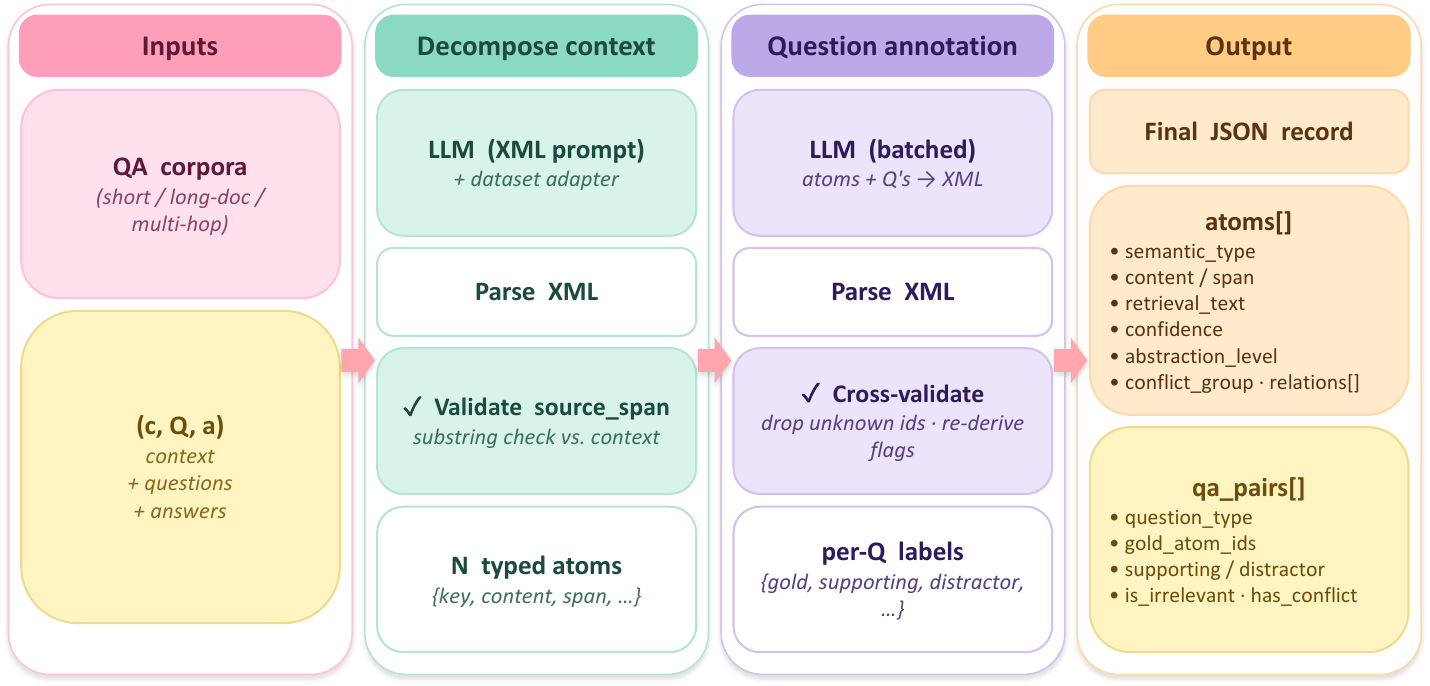}
\caption{\textbf{Atom annotation pipeline.}
An XML-driven annotator first decomposes each context into typed
atoms, guided by a dataset-specific adapter; we then parse the
output and verify every \texttt{source\_span} against the original
context. A second batched annotator call labels each question over
the validated atom bank with per-question role fields, after which
we apply deterministic cross-validation (e.g., dropping unknown
\texttt{atom\_id}s). The merged record is serialised as JSON with
\texttt{atoms[]} and \texttt{qa\_pairs[]} arrays; the full field
schema is given in Table~\ref{tab:annotation-schema}.}
\vspace{-0.3cm}
\label{fig:annotation-pipeline}
\end{figure*}

\begin{table*}[t]
\centering
\small
\setlength{\tabcolsep}{6pt}
\begin{tabular}{@{}l p{11.5cm}@{}}
\toprule
\textbf{Field} & \textbf{Description} \\
\midrule
\multicolumn{2}{@{}l}{\textit{Atom-level}} \\
\midrule
\texttt{atom\_id}            & Stable identifier (\texttt{atom\_0},
                               \texttt{atom\_1}, \dots) used for
                               cross-referencing within the sample. \\
\texttt{semantic\_type}      & One of the five types
                               \{\texttt{fact\_claim},
                               \texttt{entity\_attribute},
                               \texttt{event\_relation},
                               \texttt{process\_step},
                               \texttt{evidence\_fragment}\}. \\
\texttt{content}             & Self-contained natural-language
                               statement of the atom. \\
\texttt{source\_span}        & Verbatim contiguous substring of the
                               original context, validated post-hoc
                               (\S\ref{app:annotation}). \\
\texttt{retrieval\_text}     & Query-friendly form (entity, attribute,
                               and key qualifiers), used as the
                               indexing anchor for routing. \\
\texttt{is\_answer\_bearing} & Whether the atom can directly carry an
                               answer, as opposed to only providing
                               context. \\
\texttt{abstraction\_level}  & One of \{\texttt{abstract},
                               \texttt{evidence},
                               \texttt{hybrid}\}, used by the
                               router's metadata bias. \\
\texttt{conflict\_group}     & Cluster identifier shared by mutually
                               inconsistent atoms; \texttt{null}
                               otherwise. \\
\texttt{confidence}          & Annotator's calibrated confidence in
                               $[0,1]$. \\
\texttt{relations}           & Typed edges to other atoms; one of nine
                               types (\texttt{supports},
                               \texttt{elaborates}, \texttt{causes},
                               \texttt{precedes}, \texttt{follows},
                               \texttt{contradicts},
                               \texttt{same\_entity},
                               \texttt{same\_event},
                               \texttt{part\_of}). \\
\midrule
\multicolumn{2}{@{}l}{\textit{Question-level}} \\
\midrule
\texttt{question\_type}            & One of ten types
                                     (\texttt{single\_hop},
                                     \texttt{multi\_hop},
                                     \texttt{comparison},
                                     \texttt{temporal},
                                     \texttt{causal},
                                     \texttt{procedural},
                                     \texttt{aggregation},
                                     \texttt{field\_lookup},
                                     \texttt{evidence\_grounded},
                                     \texttt{irrelevant}). \\
\texttt{gold\_atom\_ids}           & Minimal set of atoms strictly
                                     required to answer the
                                     question. \\
\texttt{supporting\_atom\_ids}     & Atoms that aid reasoning but are
                                     not strictly required. \\
\texttt{distractor\_atom\_ids}     & Hard negatives: atoms a router
                                     would most plausibly
                                     mis-activate. \\
\texttt{relevant\_atom\_ids}       & Convenience union \texttt{gold}
                                     $\cup$ \texttt{supporting},
                                     re-derived after validation. \\
\texttt{is\_irrelevant}            & True iff the document does not
                                     contain the answer; forces
                                     \texttt{gold\_atom\_ids}
                                     $=\emptyset$. \\
\texttt{has\_conflict}             & True iff any \texttt{gold} or
                                     \texttt{supporting} atom shares a
                                     \texttt{conflict\_group} with
                                     another; re-derived from the
                                     atoms. \\
\texttt{is\_generated\_irrelevant} & True iff the question was
                                     synthesised by the
                                     irrelevant-probe generator rather
                                     than drawn from the source
                                     corpus. \\
\bottomrule
\end{tabular}
\caption{\textbf{Output schema produced by our annotation pipeline.}
Atom-level fields are emitted once per atom during context
decomposition; question-level fields are emitted once per question
during question-to-atom annotation.}
\vspace{-0.3cm}
\label{tab:annotation-schema}
\end{table*}

\section{Atomization Pipeline}
\label{app:annotation}
This section provides the implementation details of the annotation
pipeline introduced in \S\ref{sec:method-data}, focusing on three
aspects deferred from the main text: (i) the prompt and parsing
format, (ii) the deterministic post-processing rules applied after
each annotator call, and (iii) the failure-handling policy.  The operational flow is illustrated in Fig.~\ref{fig:annotation-pipeline}, and output schema is summarized in Table~\ref{tab:annotation-schema}.

\paragraph{Annotator and prompt format.}
All atom annotations are generated using MiniMax-M2.5~\citep{MinimaxM25} with structured XML prompts. Every post-processing step
downstream of the annotator is deterministic, so a fixed annotator
output always yields the same final record. We adopt XML rather
than JSON for both input and output for three reasons. First, XML
is robust to partial output: a single malformed tag does not
invalidate the surrounding document, whereas a single JSON syntax
error typically discards the entire record. Second, XML supports
per-tag validation, so we can selectively retain or repair
well-formed sub-trees even when other parts of the response are
corrupted. Third, XML degrades gracefully under mid-response
truncation, preserving every well-formed tag emitted before the
cutoff. On a held-out calibration set, XML parsing recovers a
larger fraction of usable samples than JSON parsing under matched
decoding budgets.

\subsection{Context Decomposition}
Given a context $c$, the annotator emits an XML-tagged list of
atoms, each a self-contained, minimally redundant unit of
knowledge. The five semantic types and eight metadata fields
attached to every atom are introduced in \S\ref{sec:method-data}
and tabulated in Table~\ref{tab:annotation-schema}. In
addition to these per-atom fields, atoms are linked to one another
through nine typed \emph{relations} that encode bridge entities,
supporting evidence, contradictions, and procedural ordering.

The decomposition prompt consists of a base system prompt and a
dataset-specific adapter that captures the structural priors of the
source corpus. For multi-hop corpora, the adapter instructs the
annotator to surface bridge entities and to link parallel mentions
of the same referent via \texttt{same\_entity} relations. For
numeric-reasoning corpora, it requires every comparable quantity to
be placed in its own atom whose \texttt{source\_span} is the
verbatim character sequence from the document.

\paragraph{Source-span validation.}
Every atom must carry a \texttt{source\_span} field: a contiguous
substring of the original context that grounds the atom. Immediately
after parsing, we run a substring search (optionally fuzzy with
whitespace normalisation) of each span against $c$. Atoms whose span
cannot be located receive a \texttt{span\_valid}=\textsc{false} flag
and are surfaced to the downstream filter. This single guard rule
eliminates the most common annotator hallucination, fabricated
quotations, and we find it sufficient in practice without a separate
verification call to a second annotator.

\subsection{Irrelevant Probe Generation}
\label{app:irrelevant-probes}
To support the irrelevant-suppression objective
$\mathcal{L}_{\text{irrel}}$ (\S\ref{sec:method-training}), we
augment each context with synthesised \emph{irrelevant probes}:
questions that mention entities or topics from the context but
cannot be answered from it. A separate prompt, conditioned on the
same context, asks the annotator to produce $n$ such
plausible-but-unanswerable questions. Generated probes carry
\texttt{is\_generated\_irrelevant=true}, and the subsequent
question-to-atom annotation step forces their
\texttt{gold\_atom\_ids} to be empty.

\subsection{Question-to-Atom Annotation}
Given the validated atom bank and the augmented question list $Q$
(the original questions plus the irrelevant probes generated above),
the annotator labels each question with the role fields listed in
Table~\ref{tab:annotation-schema} (bottom half). The two most
consequential roles are \texttt{gold\_atom\_ids}, the minimal atom
set strictly necessary to answer the question, and
\texttt{distractor\_atom\_ids}, the atoms a router would most
plausibly mis-activate. The dataset-specific adapter again shapes
what counts as a hard distractor: on multi-hop corpora, atoms that
bind the right attribute to the wrong entity; on field-lookup
corpora, atoms with the same entity but a different field, or the
same field but a different entity.

\paragraph{Cross-validation.}
Rather than accepting the annotator's question-level output as is,
we apply three deterministic checks: (i) drop any \texttt{atom\_id}
that does not appear in the validated atom bank; (ii) recompute
\texttt{relevant\_atom\_ids}$=$\texttt{gold}$\cup$\texttt{supporting}
from the surviving lists; and (iii) re-derive \texttt{has\_conflict}
from the atoms' own \texttt{conflict\_group} fields rather than the
annotator-emitted flag. Whenever the annotator marks a question
\texttt{is\_irrelevant}, we additionally clear its
\texttt{gold\_atom\_ids} and set \texttt{question\_type=irrelevant}
so that the two flags remain consistent.

\subsection{Failure Handling and Quality Logging}
We handle annotator failures by graceful degradation. A failure during
context decomposition discards the whole sample, as all downstream
stages need a valid atom bank. A failure during question-to-atom
annotation retains the atom bank but emits empty question annotations,
leaving an atom-only training record. A failure during irrelevant-probe
generation is logged and ignored, since the probes are auxiliary.
Per-sample warnings (e.g., invalid semantic types, dropped relation
targets, span mismatches, unknown atom IDs) are accumulated into an
\texttt{annotation\_meta} field, letting downstream consumers filter on
annotation quality without re-running the annotator.

\section{Implementation Details}
\label{app:impl}

\paragraph{Hardware and precision.}
All experiments are conducted on a cluster of 8 NVIDIA A100 80\,GB GPUs
interconnected via NVLink, using PyTorch 2.1 with DeepSpeed ZeRO-2 for
distributed training. We use FlashAttention-2 for efficient attention
throughout both training and inference. Both base models are kept
entirely frozen in bfloat16, while all trainable modules, including the
atom encoder projection head, the memory compiler trunk and heads, and
the query router, are maintained in float32 for optimization stability.

\paragraph{Memory compiler and router.}
For the memory compiler, we set the latent dimension $d=256$ and use
LoRA rank $r=8$ with scaling factor $\alpha=16$ and zero dropout,
targeting the query, value, output, and down-projection modules in the
last $n_L=4$ decoder layers. The shared text encoder extracts hidden
states from the first $L_{\text{enc}}=4$ layers of the frozen base LLM
for both models. The two-stage router uses high-recall retrieval with
$K_1=32$ followed by a final selection of $K=8$ atoms. The learned
4-dimensional metadata bias is enabled by default, and the 
second-stage reranker follows the configuration in
Table~\ref{tab:hyperparams}.

\paragraph{Optimization.}
We optimize with AdamW ($\beta_1=0.9$, $\beta_2=0.999$) at a learning
rate of $1\times10^{-4}$ on Gemma-2-2B-It and $8\times10^{-5}$ on
Qwen3-4B-Instruct-2507, with cosine scheduling, 500 warm-up steps,
weight decay $0.01$, and gradient clipping at $1.0$. The effective
global batch size is 128, and the distillation temperature is
$\tau=1.5$.

\paragraph{Data and decoding.}
The maximum context length is 8192 tokens, with atom texts truncated to 96 tokens and answers to 128 tokens; each sample contains at most 64 atoms and 8 questions. Evaluation uses greedy decoding with up to 128 new tokens. The full curriculum schedule (loss-weight transitions, top-$K$ ramps, and stage boundaries) is given in \S\ref{app:curriculum}, and the complete per-LLM hyperparameter configuration is listed in Table~\ref{tab:hyperparams}.

\section{Curriculum}
\label{app:curriculum}

Training proceeds through a four-stage curriculum that gradually increases task difficulty and activates losses in dependency order. Each stage is specified by a step range, the top-$K$ atom budget passed to the router, and the set of active losses.

\begin{enumerate}[label=\textbf{Stage \arabic*.},leftmargin=*,itemsep=2pt]
    \item \textbf{Single-atom memory} (steps 0--2000, $K{=}1$).
    The router selects exactly one atom per query. Only the core
    losses are active: $\mathcal{L}_{\text{LM}}$,
    $\mathcal{L}_{\text{distill}}$, $\mathcal{L}_{\text{route}}$,
    $\mathcal{L}_{\text{irrel}}$, and $\mathcal{L}_{\text{protect}}$.
    \item \textbf{Multi-atom composition} (steps 2000--6000, $K{=}4$).
    The router selects four atoms;
    $\mathcal{L}_{\text{composition}}$ and
    $\mathcal{L}_{\text{sparse}}$ become active.
    \item \textbf{Robustness} (steps 6000--9000, $K{=}8$).
    The router selects eight atoms; the auxiliary robustness losses
    $\mathcal{L}_{\text{conflict}}$ and
    $\mathcal{L}_{\text{confidence}}$ are added, and the weight on
    $\mathcal{L}_{\text{irrel}}$ is increased.
    \item \textbf{Full composition} (steps 9000$+$, $K{=}8$).
    All losses remain active and the weight on
    $\mathcal{L}_{\text{sparse}}$ is strengthened to push the gate
    density closer to the target $\rho$.
\end{enumerate}

Training runs for up to 50 epochs with early stopping on the validation language-modeling loss.

\begin{table}[ht]
\centering
\small
\setlength{\tabcolsep}{8pt}
\renewcommand{\arraystretch}{1.1}
\begin{tabular}{@{}l cc@{}}
\toprule
Dataset & Base+ctx & \ours{}+ctx \\
\midrule
\multicolumn{3}{c}{\textit{F1}} \\
\midrule
2WikiMultiHopQA         & 37.69 & \textbf{40.44} \\
DROP                    & 42.61 & \textbf{49.13} \\
QASPER                  & \textbf{48.62} & 46.90 \\
ROPES                   & 71.90 & \textbf{73.07} \\
SQuAD                   & 86.35 & \textbf{90.10} \\
\midrule
LongBench-2WikiMQA      & 32.27 & \textbf{33.57} \\
LongBench-HotpotQA      & \textbf{41.00} & 39.07 \\
LongBench-MFQA-en       & 39.41 & \textbf{46.55} \\
LongBench-MFQA-zh       & \phantom{0}7.73 & \textbf{12.91} \\
LongBench-MuSiQue       & 21.62 & \textbf{30.73} \\
LongBench-NarrativeQA   & 22.67 & \textbf{27.23} \\
LongBench-QASPER        & 37.05 & \textbf{41.46} \\
LongBench-TriviaQA      & \textbf{86.67} & 78.83 \\
\midrule
\textit{Overall}        & 48.79 & \textbf{51.05} \\
\midrule
\multicolumn{3}{c}{\textit{ROUGE-L}} \\
\midrule
2WikiMultiHopQA         & 37.69 & \textbf{40.44} \\
DROP                    & 42.61 & \textbf{49.13} \\
QASPER                  & \textbf{47.26} & 45.69 \\
ROPES                   & 71.90 & \textbf{73.07} \\
SQuAD                   & 86.20 & \textbf{90.02} \\
\midrule
LongBench-2WikiMQA      & 32.27 & \textbf{33.57} \\
LongBench-HotpotQA      & \textbf{40.78} & 38.99 \\
LongBench-MFQA-en       & 38.31 & \textbf{45.60} \\
LongBench-MFQA-zh       & \phantom{0}7.73 & \textbf{12.71} \\
LongBench-MuSiQue       & 21.62 & \textbf{30.73} \\
LongBench-NarrativeQA   & 22.37 & \textbf{27.08} \\
LongBench-QASPER        & 35.58 & \textbf{40.29} \\
LongBench-TriviaQA      & \textbf{86.67} & 78.83 \\
\midrule
\textit{Overall}        & 48.43 & \textbf{50.74} \\
\bottomrule
\end{tabular}
\caption{\textbf{In-context comparison on Gemma-2-2B-It.} Both models receive the full document in the prompt: the base model (Base+ctx) versus \ours{}+ctx, which additionally conditions on its compiled atom memory. We report F1 and ROUGE-L (\%), respectively; the better score in each row is in \textbf{bold}.}
\vspace{-0.5cm}
\label{tab:gemma-base-ctx-vs-ours-ctx}
\end{table}

\section{Atom Memory Complements In-Context Access}
\label{app:ctx-complement}
\ours{} is designed to answer without the source document in the prompt. Here we test whether its compiled atom memory still helps when the document \emph{is} available, i.e.\ whether the atoms add signal beyond what the model can already read. We compare the base LLM with the document in context (Base+ctx) against \ours{} given the same document plus its atom memory (\ours{}+ctx). All other settings match the main experiments.

As shown in Table~\ref{tab:gemma-base-ctx-vs-ours-ctx} on Gemma-2-2B-It, adding the atom memory raises Overall F1 from $48.79$ to
$51.05$ ($+2.26$), with gains on most datasets and the same trend under ROUGE-L. The improvements are largest on aggregation and multi-hop tasks (LongBench-MuSiQue $+9.11$, LongBench-MFQA-en $+7.14$, DROP $+6.52$), where a structured, pre-decomposed view of the document is most useful. The atom memory does not help everywhere: on a few datasets (e.g.,  TriviaQA, a short-answer factoid task whose answers are already easy to locate in the context) it brings little or no gain. Overall, these results suggest that the atom memory is largely complementary to in-context access, adding useful structure in the cases where reading raw text alone falls short.

\section{Atomization Example}
\label{app:case-study}
To illustrate the pipeline of \S\ref{app:annotation},
Fig.~\ref{fig:atomization-example-part1}  walks through an uncurated, randomly selected atomization example from 2WikiMultiHopQA  benchmark \citep{ho2020constructing}: Part~1 shows the raw 10-passage document;
Parts~2--3 list the 17 extracted atoms with their metadata; and
Part~3 adds the questions with their
gold/supporting/distractor labels.

\definecolor{TypeEntityAttr}{HTML}{A8E6CF}
\definecolor{TypeFactClaim}{HTML}{FFC2C8}
\definecolor{TypeEventRel}{HTML}{FFE699}
\definecolor{Anchor}{HTML}{FFF7CA}
\definecolor{GoldChip}{HTML}{D97706}
\definecolor{SupportChip}{HTML}{2563EB}
\definecolor{DistractChip}{HTML}{7C3AED}
\definecolor{QMultiHop}{HTML}{F97316}
\definecolor{QIrrel}{HTML}{DC2626}
\definecolor{WarnCol}{HTML}{B91C1C}

\sethlcolor{Anchor}   % soul highlight uses Anchor as background

\newcommand{\typechip}[2]{\fcolorbox{black!30}{#1}{\strut\scriptsize\bfseries\,#2\,}}
\newcommand{\aid}[1]{\texttt{a$_{\mathrm{#1}}$}}
\newcommand{\yes}{\textcolor{green!55!black}{$\checkmark$}}
\newcommand{\no}{\textcolor{red!70!black}{$\boldsymbol{\times}$}}
\newcommand{\rel}[2]{\,$\rightarrow$\,\aid{#1}\,(\textit{#2})}

\newcommand{\atomcard}[4]{%
  \begin{tcolorbox}[colback=white, colframe=#2!70!black, sharp corners,
    boxsep=2pt, left=4pt, right=4pt, top=2pt, bottom=2pt, before skip=2pt, after skip=2pt]
    \scriptsize
    \textbf{\aid{#1}}\, \typechip{#2}{#3}\par #4
  \end{tcolorbox}}

% ===========================================================================
%  Part 1 of 3 :  (a) FULL input document
% ===========================================================================
\begin{figure*}[ht]
\centering
\footnotesize
\setlength{\parskip}{2pt}

\begin{tcolorbox}[
  enhanced, colback=white, colframe=black!25,
  sharp corners, boxsep=3pt, left=5pt, right=5pt, top=5pt, bottom=5pt,
  title={Atomization Example (Part 1 of 3) ---
         2WikiMultiHopQA, id \texttt{2wiki\_000b2e3d098711ebbdb0ac1f6bf848b6},
         type \emph{comparison}, 17 atoms / 3 questions},
  fonttitle=\small\bfseries,
]

\textbf{(a) Input document (10 Wikipedia stub passages, full text)}\hfill
{\scriptsize passages relevant to $q_0$ and $q_1$ are highlighted in
\colorbox{Anchor}{\strut soft yellow}}
\par\smallskip
\begin{quote}\sloppy\small

\hl{\textit{Passage [Telephone numbers in Ascension Island]:} Country Code:+ 247 <br> International Call Prefix: 00 Ascension Island does not share the same country code( +290) with the rest of St Helena.}

\textit{Passage [Satellite tournament]:} A satellite tournament is either a minor tournament or event on a competitive sporting tour or one of a group of such tournaments that form a series played in the same country or region.

\hl{\textit{Passage [Qaserdalu]:} Qaserdalu( also Romanized as Qaṣerdālū) is a village in Tolbozan Rural District, Golgir District, Masjed Soleyman County, Khuzestan Province, Iran. At the 2006 census, its population was 97, in 23 families.}

\textit{Passage [Lubnowy]:} Lubnowy is part of the name of two villages, both located in Gmina Susz, within Iława County, Warmian- Masurian Voivodeship, Poland:

\textit{Passage [Radzice]:} Radzice is part of the name of two villages, both located in Gmina Drzewica, within Opoczno County, Łódź Voivodeship, Poland:

\textit{Passage [Limestone Coast]:} The Limestone Coast is a name used since the early twenty- first century for a South Australian government region located in the south east of South Australia which immediately adjoins the continental coastline and the Victorian border. The name is also used for a tourist region and a wine zone both located in the same part of South Australia.

\textit{Passage [Tunstall, Virginia]:} Tunstall is an unincorporated community in New Kent County, Virginia, United States. Foster's Castle and Hampstead, both located in Tunstall, are listed on the National Register of Historic Places.

\hl{\textit{Passage [Quyujoq]:} Quyujoq( also Romanized as Qūyūjoq; also known as Kūjīkh and Qūjūq) is a village in Anzal -e Jonubi Rural District, Anzal District, Urmia County, West Azerbaijan Province, Iran. At the 2006 census, its population was 93, in 14 families.}

\textit{Passage [Jawty]:} Jawty is part of the name of two villages, both located in Gmina Susz, within Iława County, Warmian- Masurian Voivodeship, Poland:

\textit{Passage [Greenbury Point Light]:} Greenbury Point Light was the name of two lighthouses in the Chesapeake Bay, both located at the mouth of the Severn River in Annapolis, Maryland.
\end{quote}

\end{tcolorbox}

\caption{%
\textbf{An Atomization Example (Part 1 of 3).}}
\label{fig:atomization-example-part1}
\end{figure*}

% ===========================================================================
%  Part 2 of 3 :  atoms a_0 .. a_9 
% ===========================================================================
\begin{figure*}[!t]
\ContinuedFloat
\centering
\footnotesize
\setlength{\parskip}{2pt}

\begin{tcolorbox}[
  enhanced, colback=white, colframe=black!25,
  sharp corners, boxsep=3pt, left=5pt, right=5pt, top=5pt, bottom=5pt,
  title={Atomization Example (Part 2 of 3) --- 17 annotated atoms, first half ($a_0$--$a_9$)},
  fonttitle=\small\bfseries,
]

\textbf{(b) 17 annotated atoms --- first half ($a_0$--$a_9$)} \hfill
{\scriptsize semantic types in this example:}
\typechip{TypeEntityAttr}{entity\_attribute}\;
\typechip{TypeFactClaim}{fact\_claim}

\smallskip
\noindent
\begin{minipage}[t]{0.49\textwidth}
\atomcard{0}{TypeEntityAttr}{entity\_attribute}{%
conf=0.95 \,|\, span [49,67]\,\yes \,|\, evidence \,|\, answer-bearing\,\yes\par
\textbf{content:} Ascension Island has country code +247.\par
\textbf{source\_span:} \textit{``Country Code:+ 247''}\par
\textbf{retrieval:} \texttt{Ascension Island country code}\par
\textbf{relations:} (none)\quad\textbf{conflict\_group:} (none)}

\atomcard{2}{TypeFactClaim}{fact\_claim}{%
conf=0.95 \,|\, span [103,191]\,\yes \,|\, evidence \,|\, answer-bearing\,\yes\par
\textbf{content:} Ascension Island does not share the same country code (+290) with the rest of St Helena.\par
\textbf{source\_span:} \textit{``Ascension Island does not share the same country code( +290) with the rest of St Helena.''}\par
\textbf{retrieval:} \texttt{Ascension Island different country code from St Helena}\par
\textbf{relations:} (none)\quad\textbf{conflict\_group:} (none)}

\atomcard{4}{TypeEntityAttr}{entity\_attribute}{%
conf=0.95 \,|\, span [434,581]\,\yes \,|\, evidence \,|\, answer-bearing\,\yes\par
\textbf{content:} Qaserdalu is a village in Tolbozan Rural District, Golgir District, Masjed Soleyman County, Khuzestan Province, Iran.\par
\textbf{source\_span:} \textit{``Qaserdalu( also Romanized as Qaṣerdālū) is a village in Tolbozan Rural District, Golgir District, Masjed Soleyman County, Khuzestan Province, Iran.''}\par
\textbf{retrieval:} \texttt{Qaserdalu location}\par
\textbf{relations:} (none)\quad\textbf{conflict\_group:} (none)}

\atomcard{6}{TypeEntityAttr}{entity\_attribute}{%
conf=0.95 \,|\, span [661,793]\,\yes \,|\, evidence \,|\, answer-bearing\,\yes\par
\textbf{content:} Lubnowy is part of the name of two villages in Gmina Susz, Iława County, Warmian-Masurian Voivodeship, Poland.\par
\textbf{source\_span:} \textit{``Lubnowy is part of the name of two villages, both located in Gmina Susz, within Iława County, Warmian- Masurian Voivodeship, Poland:''}\par
\textbf{retrieval:} \texttt{Lubnowy villages location}\par
\textbf{relations:} (none)\quad\textbf{conflict\_group:} (none)}

\atomcard{8}{TypeEntityAttr}{entity\_attribute}{%
conf=0.95 \,|\, span [968,1126]\,\yes \,|\, evidence \,|\, answer-bearing\,\yes\par
\textbf{content:} The Limestone Coast is a South Australian government region in the south east of South Australia.\par
\textbf{source\_span:} \textit{``The Limestone Coast is a name used since the early twenty- first century for a South Australian government region located in the south east of South Australia''}\par
\textbf{retrieval:} \texttt{Limestone Coast location and name origin}\par
\textbf{relations:} (none)\quad\textbf{conflict\_group:} (none)}
\end{minipage}\hfill
\begin{minipage}[t]{0.49\textwidth}
\atomcard{1}{TypeEntityAttr}{entity\_attribute}{%
conf=0.95 \,|\, span [73,102]\,\yes \,|\, evidence \,|\, answer-bearing\,\yes\par
\textbf{content:} International Call Prefix for Ascension Island is 00.\par
\textbf{source\_span:} \textit{``International Call Prefix: 00''}\par
\textbf{retrieval:} \texttt{Ascension Island international call prefix}\par
\textbf{relations:} (none)\quad\textbf{conflict\_group:} (none)}

\atomcard{3}{TypeFactClaim}{fact\_claim}{%
conf=0.90 \,|\, span [225,411]\,\yes \,|\, \textbf{abstract} \,|\, answer-bearing\,\yes\par
\textbf{content:} A satellite tournament is a minor tournament on a competitive sporting tour or one of a group of tournaments in the same country or region.\par
\textbf{source\_span:} \textit{``A satellite tournament is either a minor tournament or event on a competitive sporting tour or one of a group of such tournaments that form a series played in the same country or region.''}\par
\textbf{retrieval:} \texttt{satellite tournament definition}\par
\textbf{relations:} (none)\quad\textbf{conflict\_group:} (none)}

\atomcard{5}{TypeEntityAttr}{entity\_attribute}{%
conf=0.95 \,|\, span [582,640]\,\yes \,|\, evidence \,|\, answer-bearing\,\yes\par
\textbf{content:} Qaserdalu had population of 97 in 23 families in 2006.\par
\textbf{source\_span:} \textit{``At the 2006 census, its population was 97, in 23 families.''}\par
\textbf{retrieval:} \texttt{Qaserdalu 2006 census population}\par
\textbf{relations:} (none)\quad\textbf{conflict\_group:} (none)}

\atomcard{7}{TypeEntityAttr}{entity\_attribute}{%
conf=0.95 \,|\, span [814,939]\,\yes \,|\, evidence \,|\, answer-bearing\,\yes\par
\textbf{content:} Radzice is part of the name of two villages in Gmina Drzewica, Opoczno County, Łódź Voivodeship, Poland.\par
\textbf{source\_span:} \textit{``Radzice is part of the name of two villages, both located in Gmina Drzewica, within Opoczno County, Łódź Voivodeship, Poland:''}\par
\textbf{retrieval:} \texttt{Radzice villages location}\par
\textbf{relations:} (none)\quad\textbf{conflict\_group:} (none)}

\atomcard{9}{TypeEntityAttr}{entity\_attribute}{%
conf=0.95 \,|\, span [1127,1204]\,\yes \,|\, evidence \,|\, answer-bearing\,\yes\par
\textbf{content:} The Limestone Coast adjoins the continental coastline and the Victorian border.\par
\textbf{source\_span:} \textit{``which immediately adjoins the continental coastline and the Victorian border.''}\par
\textbf{retrieval:} \texttt{Limestone Coast geographical boundaries}\par
\textbf{relations:} (none)\quad\textbf{conflict\_group:} (none)}
\end{minipage}

\end{tcolorbox}

\caption{%
\textbf{An Atomization Example (continued, Part 2 of 3).}}
\label{fig:atomization-example-part2}
\end{figure*}

% ===========================================================================
%  Part 3 of 3 :  atoms a_10 .. a_16  +  3 questions  +  annotation_meta
% ===========================================================================
\begin{figure*}[!t]
\ContinuedFloat
\centering
\footnotesize
\setlength{\parskip}{2pt}

\begin{tcolorbox}[
  enhanced, colback=white, colframe=black!25,
  sharp corners, boxsep=3pt, left=5pt, right=5pt, top=5pt, bottom=5pt,
  title={Atomization Example (Part 3 of 3) ---
         atoms $a_{10}$--$a_{16}$, 3 annotated questions, annotation meta},
  fonttitle=\small\bfseries,
]

\textbf{(b) 17 annotated atoms --- second half ($a_{10}$--$a_{16}$)}\hfill
{\scriptsize types:}
\typechip{TypeEntityAttr}{entity\_attribute}\;
\typechip{TypeFactClaim}{fact\_claim}

\smallskip
\noindent
\begin{minipage}[t]{0.49\textwidth}
\atomcard{10}{TypeEntityAttr}{entity\_attribute}{%
conf=0.95 \,|\, span [1205,1313]\,\yes \,|\, evidence \,|\, answer-bearing\,\yes\par
\textbf{content:} The Limestone Coast name is also used for a tourist region and wine zone in the same part of South Australia.\par
\textbf{source\_span:} \textit{``The name is also used for a tourist region and a wine zone both located in the same part of South Australia.''}\par
\textbf{retrieval:} \texttt{Limestone Coast tourist and wine zone usage}\par
\textbf{relations:} (none)\quad\textbf{conflict\_group:} (none)}

\atomcard{12}{TypeEntityAttr}{entity\_attribute}{%
conf=0.95 \,|\, span [1430,1542]\,\yes \,|\, evidence \,|\, answer-bearing\,\yes\par
\textbf{content:} Foster's Castle and Hampstead are both located in Tunstall, Virginia and listed on the National Register of Historic Places.\par
\textbf{source\_span:} \textit{``Foster's Castle and Hampstead, both located in Tunstall, are listed on the National Register of Historic Places.''}\par
\textbf{retrieval:} \texttt{Historic places in Tunstall Virginia}\par
\textbf{relations:} (none)\quad\textbf{conflict\_group:} (none)}

\atomcard{14}{TypeEntityAttr}{entity\_attribute}{%
conf=0.95 \,|\, span [1741,1799]\,\yes \,|\, evidence \,|\, answer-bearing\,\yes\par
\textbf{content:} Quyujoq had population of 93 in 14 families in 2006.\par
\textbf{source\_span:} \textit{``At the 2006 census, its population was 93, in 14 families.''}\par
\textbf{retrieval:} \texttt{Quyujoq 2006 census population}\par
\textbf{relations:} (none)\quad\textbf{conflict\_group:} (none)}

\atomcard{16}{TypeEntityAttr}{entity\_attribute}{%
conf=0.95 \,|\, span [1983,2129]\,\yes \,|\, evidence \,|\, answer-bearing\,\yes\par
\textbf{content:} Greenbury Point Light refers to two lighthouses at the mouth of the Severn River in Annapolis, Maryland.\par
\textbf{source\_span:} \textit{``Greenbury Point Light was the name of two lighthouses in the Chesapeake Bay, both located at the mouth of the Severn River in Annapolis, Maryland.''}\par
\textbf{retrieval:} \texttt{Greenbury Point Light location}\par
\textbf{relations:} (none)\quad\textbf{conflict\_group:} (none)}
\end{minipage}\hfill
\begin{minipage}[t]{0.49\textwidth}
\atomcard{11}{TypeEntityAttr}{entity\_attribute}{%
conf=0.95 \,|\, span [1345,1429]\,\yes \,|\, evidence \,|\, answer-bearing\,\yes\par
\textbf{content:} Tunstall is an unincorporated community in New Kent County, Virginia, United States.\par
\textbf{source\_span:} \textit{``Tunstall is an unincorporated community in New Kent County, Virginia, United States.''}\par
\textbf{retrieval:} \texttt{Tunstall Virginia location}\par
\textbf{relations:} (none)\quad\textbf{conflict\_group:} (none)}

\atomcard{13}{TypeEntityAttr}{entity\_attribute}{%
conf=0.95 \,|\, span [1563,1740]\,\yes \,|\, evidence \,|\, answer-bearing\,\yes\par
\textbf{content:} Quyujoq is a village in Anzal-e Jonubi Rural District, Anzal District, Urmia County, West Azerbaijan Province, Iran.\par
\textbf{source\_span:} \textit{``Quyujoq( also Romanized as Qūyūjoq; also known as Kūjīkh and Qūjūq) is a village in Anzal -e Jonubi Rural District, Anzal District, Urmia County, West Azerbaijan Province, Iran.''}\par
\textbf{retrieval:} \texttt{Quyujoq location}\par
\textbf{relations:} (none)\quad\textbf{conflict\_group:} (none)}

\atomcard{15}{TypeEntityAttr}{entity\_attribute}{%
conf=0.95 \,|\, span [1818,1948]\,\yes \,|\, evidence \,|\, answer-bearing\,\yes\par
\textbf{content:} Jawty is part of the name of two villages in Gmina Susz, Iława County, Warmian-Masurian Voivodeship, Poland.\par
\textbf{source\_span:} \textit{``Jawty is part of the name of two villages, both located in Gmina Susz, within Iława County, Warmian- Masurian Voivodeship, Poland:''}\par
\textbf{retrieval:} \texttt{Jawty villages location}\par
\textbf{relations:} (none)\quad\textbf{conflict\_group:} (none)}
\end{minipage}

% ---------------------------------------------------------------------------
% (c) Questions
% ---------------------------------------------------------------------------
\smallskip
\textbf{(c) 3 annotated questions}\hfill
{\scriptsize legend:
\textcolor{GoldChip}{\textbf{gold}}\;
\textcolor{SupportChip}{\textbf{support}}\;
\textcolor{DistractChip}{\textbf{distractor}}}

\smallskip
\begin{tcolorbox}[colback=white, colframe=QMultiHop!80!black, sharp corners,
  boxsep=2pt, left=4pt, right=4pt, top=2pt, bottom=2pt, before skip=2pt, after skip=2pt]
\scriptsize
\textbf{\texttt{q\_0}}\, \typechip{QMultiHop!30}{comparison} \,|\,
\textbf{is\_irrelevant:}\,\no \,|\, \textbf{has\_conflict:}\,\no \,|\,
\textbf{is\_generated\_irrelevant:}\,\no\par
\textbf{Q:} Are Quyujoq and Qaserdalu both located in the same country?\par
\textbf{\textcolor{GoldChip}{gold\_atom\_ids}}: \aid{4},\,\aid{13}\quad
\textbf{\textcolor{SupportChip}{supporting\_atom\_ids}}: ---\quad
\textbf{\textcolor{DistractChip}{distractor\_atom\_ids}}: \aid{6},\,\aid{7},\,\aid{15}\par
\textbf{relevant\_atom\_ids:} \aid{4},\,\aid{13}\quad
\textbf{annotation\_warnings:} (none)
\end{tcolorbox}

\begin{tcolorbox}[colback=white, colframe=QMultiHop!80!black, sharp corners,
  boxsep=2pt, left=4pt, right=4pt, top=2pt, bottom=2pt, before skip=2pt, after skip=2pt]
\scriptsize
\textbf{\texttt{q\_1}}\, \typechip{QMultiHop!30}{field\_lookup} \,|\,
\textbf{is\_irrelevant:}\,\no \,|\, \textbf{has\_conflict:}\,\no \,|\,
\textbf{is\_generated\_irrelevant:}\,\yes \;(\emph{generated as decoy, then rescued by Stage-2})\par
\textbf{Q:} What is the local area code used for telephone numbers within Ascension Island?\par
\textbf{\textcolor{GoldChip}{gold\_atom\_ids}}: \aid{0}\quad
\textbf{\textcolor{SupportChip}{supporting\_atom\_ids}}: \aid{1}\quad
\textbf{\textcolor{DistractChip}{distractor\_atom\_ids}}: \aid{2}\par
\textbf{relevant\_atom\_ids:} \aid{0},\,\aid{1}\quad
\textbf{annotation\_warnings:} (none)
\end{tcolorbox}

\begin{tcolorbox}[colback=white, colframe=QIrrel!80!black, sharp corners,
  boxsep=2pt, left=4pt, right=4pt, top=2pt, bottom=2pt, before skip=2pt, after skip=2pt]
\scriptsize
\textbf{\texttt{q\_2}}\, \typechip{QIrrel!30}{irrelevant} \,|\,
\textbf{is\_irrelevant:}\,\yes \,|\, \textbf{has\_conflict:}\,\no \,|\,
\textbf{is\_generated\_irrelevant:}\,\yes\par
\textbf{Q:} What are the names of specific wineries located in the Limestone Coast wine region?\par
\textbf{\textcolor{GoldChip}{gold\_atom\_ids}}: ---\quad
\textbf{\textcolor{SupportChip}{supporting\_atom\_ids}}: ---\quad
\textbf{\textcolor{DistractChip}{distractor\_atom\_ids}}: ---\par
\textbf{relevant\_atom\_ids:} ---\quad
\textbf{annotation\_warnings:} (none)
\end{tcolorbox}

\smallskip
{\scriptsize\textcolor{black!60}{%
\textbf{annotation\_meta:}\;
mode=\texttt{batch2step},\;
decomposition\_parse\_status=\texttt{ok},\;
question\_parse\_status=\texttt{ok},\;
num\_atoms=17,\;
num\_questions=3,\;
num\_irrelevant\_generated=2,\;
warnings=\textbf{(none)} --- all atom spans pass the substring check and no Stage-2 question is reclassified.}}

\end{tcolorbox}

\caption{%
\textbf{An Atomization Example (continued, Part 3 of 3).}}
\label{fig:atomization-example-part3}
\end{figure*}

\definecolor{HPHeader}{HTML}{E8EEF7}    % subtle blue-grey for section banners

\definecolor{HPHeader}{HTML}{E8EEF7}

\begin{table*}[ht]
\centering
\resizebox{0.9\textwidth}{!}{
\footnotesize
\setlength{\tabcolsep}{6pt}
\renewcommand{\arraystretch}{1.18}
\begin{tabular}{@{}p{4.6cm} p{5.4cm} p{5.4cm}@{}}
\toprule
\textbf{Component}
 & \textbf{Gemma-2-2B-It}
 & \textbf{Qwen3-4B-Instruct-2507} \\
\midrule
\rowcolor{HPHeader}\multicolumn{3}{@{}l}{\textit{Base model (frozen)}}\\
Identifier
  & \textbf{\texttt{google/gemma-2-2b-it}}
  & \textbf{\texttt{Qwen/Qwen3-4B-Instruct-2507}} \\
Hidden size $d$
  & \textbf{$2304$}
  & \textbf{$2560$} \\
Decoder layers
  & \textbf{$26$}
  & \textbf{$36$} \\
Attention / KV heads
  & \textbf{$8\,/\,4$}
  & \textbf{$32\,/\,8$} \\
Head dim
  & \textbf{$256$}
  & \textbf{$128$} \\
Position cap
  & \textbf{$8192$}
  & \textbf{$32768$} \\
Weight dtype
  & \texttt{bfloat16}
  & \texttt{bfloat16} \\
Attention impl.
  & FlashAttention-2
  & FlashAttention-2 \\
\midrule
\rowcolor{HPHeader}\multicolumn{3}{@{}l}{\textit{Atom encoder (shared between atom and query)}}\\
Backbone (frozen)
  & first $4$ layers of base LLM
  & first $4$ layers of base LLM \\
Encoder dtype
  & \texttt{bfloat16}
  & \texttt{bfloat16} \\
Pooling
  & mean over attended tokens
  & mean over attended tokens \\
Projection head
  & LN $\to$ Lin$(2304{\to}256)$ $\to$ GELU $\to$ Lin$(256{\to}256)$
  & LN $\to$ Lin$(2560{\to}256)$ $\to$ GELU $\to$ Lin$(256{\to}256)$ \\
Output dim
  & $256$
  & $256$ \\
\midrule
\rowcolor{HPHeader}\multicolumn{3}{@{}l}{\textit{Memory compiler}}\\
Trunk
  & LN $\to$ Lin$(256{\to}256)$ $\to$ GELU $\to$ Lin$(256{\to}256)$
  & LN $\to$ Lin$(256{\to}256)$ $\to$ GELU $\to$ Lin$(256{\to}256)$ \\
Latent / key dim
  & $256$
  & $256$ \\
LoRA rank $r$ / scale $\alpha$
  & $8\,/\,16.0$
  & $8\,/\,16.0$ \\
Memory layers
  & last $4$ blocks (indices $22$--$25$)
  & last $4$ blocks (indices $32$--$35$) \\
Target modules
  & \texttt{q\_proj}, \texttt{v\_proj}, \texttt{o\_proj}, \texttt{down\_proj}
  & \texttt{q\_proj}, \texttt{v\_proj}, \texttt{o\_proj}, \texttt{down\_proj} \\
LoRA init
  & $A\!\sim\!\mathcal{N}(0,0.02^2)$, $B{=}0$
  & $A\!\sim\!\mathcal{N}(0,0.02^2)$, $B{=}0$ \\
\midrule
\rowcolor{HPHeader}\multicolumn{3}{@{}l}{\textit{Two-stage router}}\\
Stage-1 retrieval
  & cosine over $L_2$-normalised keys
  & cosine over $L_2$-normalised keys \\
Stage-1 top-$K_1$
  & $32$
  & $32$ \\
Stage-2 reranker
  & \texttt{BAAI/bge-reranker-base} 
  & \texttt{BAAI/bge-reranker-base} \\
Stage-2 max length
  & $192$ tokens
  & $192$ tokens \\
Rerank fusion weight $\alpha_r$
  & $1.0$
  & $1.0$ \\
Metadata bias
  & 4 learned scalars
  & 4 learned scalars \\
Final selection
  & softmax-weighted top-$k$ (curriculum)
  & softmax-weighted top-$k$ (curriculum) \\
Routing temperature
  & $1.0$
  & $1.0$ \\
Composer softmax
  & on
  & on \\
\midrule
\rowcolor{HPHeader}\multicolumn{3}{@{}l}{\textit{Optimization}}\\
Optimiser
  & AdamW $(\beta_1{=}0.9,\beta_2{=}0.999)$
  & AdamW $(\beta_1{=}0.9,\beta_2{=}0.999)$ \\
Weight decay
  & $0.01$
  & $0.01$ \\
Learning rate
  & \textbf{$1.0\!\times\!10^{-4}$}
  & \textbf{$8.0\!\times\!10^{-5}$} \\
LR schedule
  & cosine
  & cosine \\
Warm-up
  & $500$ steps
  & $500$ steps \\
\midrule
\rowcolor{HPHeader}\multicolumn{3}{@{}l}{\textit{Hardware \& data loading}}\\
GPUs
  & $8\times$ NVIDIA A100 80\,GB
  & $8\times$ NVIDIA A100 80\,GB \\
Distributed
  & DDP, \texttt{find\_unused\_parameters=true}
  & DDP, \texttt{find\_unused\_parameters=true} \\
Dataloader workers / prefetch / persistent
  & $8$ per rank / $4$ / on
  & $8$ per rank / $4$ / on \\
\bottomrule
\end{tabular}
}
\caption{\textbf{Hyperparameter configuration of \ours{} on both base
LLMs.} Values are read row by row: the middle column is the gemma-2-2b-it run, the right column is the
Qwen3-4B-Instruct-2507 run.}
\label{tab:hyperparams}
\end{table*}

\end{document}